\documentclass{article} 
\usepackage{iclr2023_conference,times}

\usepackage[hidelinks]{hyperref}
\usepackage{url}
\usepackage{booktabs}
\usepackage{xcolor, colortbl}         
\usepackage{amsmath, amssymb, mathtools, bm}
\usepackage{multirow, enumitem, subcaption, floatrow, float, wrapfig}
\newfloatcommand{capbtabbox}{table}[][\FBwidth]
\usepackage{pifont}
\newcommand{\cmark}{\ding{51}}%
\newcommand{\xmark}{\ding{55}}%

\usepackage{soul}

\title{Masked Vision and Language Modeling for Multi-modal Representation Learning}

\author{Gukyeong Kwon, Zhaowei Cai, Avinash Ravichandran, \\\textbf{Erhan Bas, Rahul Bhotika, Stefano Soatto}\\
  AWS AI Labs\\
  \texttt{\{gukyeong,zhaoweic,soattos\}@amazon.com} \\
}

\iclrfinalcopy 
\begin{document}

\maketitle

\begin{abstract}
In this paper, we study how to use masked signal modeling in vision and language (V+L) representation learning. Instead of developing masked language modeling (MLM) and masked image modeling (MIM) independently, we propose to build joint masked vision and language modeling, where the masked signal of one modality is reconstructed with the help from another modality. This is motivated by the nature of image-text paired data that both of the image and the text convey almost the same information but in different formats. The masked signal reconstruction of one modality conditioned on another modality can also implicitly learn cross-modal alignment between language tokens and image patches. Our experiments on various V+L tasks show that the proposed method, along with common V+L alignment losses, achieves state-of-the-art performance in the regime of millions of pre-training data. Also, we outperforms the other competitors by a significant margin in limited data scenarios. 
\end{abstract}

\section{Introduction}\vspace{-0.15cm}
Vision and language (V+L) representation learning has gained significant attention due to the transferablility of the representations to a diverse set of downstream tasks such as zero- or few-shot visual recognition~\citep{jia2021scaling, radford2021learning, tsimpoukelli2021multimodal}, object detection~\citep{cai2022x, kamath2021mdetr}, information retrieval~\citep{li2022blip, li2021align}, and multi-modal generation~\citep{ramesh2022hierarchical, ramesh2021zero} etc. This success is mainly driven by large-scale pre-training with paired image and text data. The current V+L pre-training techniques particularly focus on the representation learning that characterizes the association between vision and language, and they are largely inspired by self-supervised learning techniques~\citep{devlin2018bert, he2020momentum} in uni-modal learning. 

\begin{figure}[t]
    \centering
    \includegraphics[width=.9\textwidth]{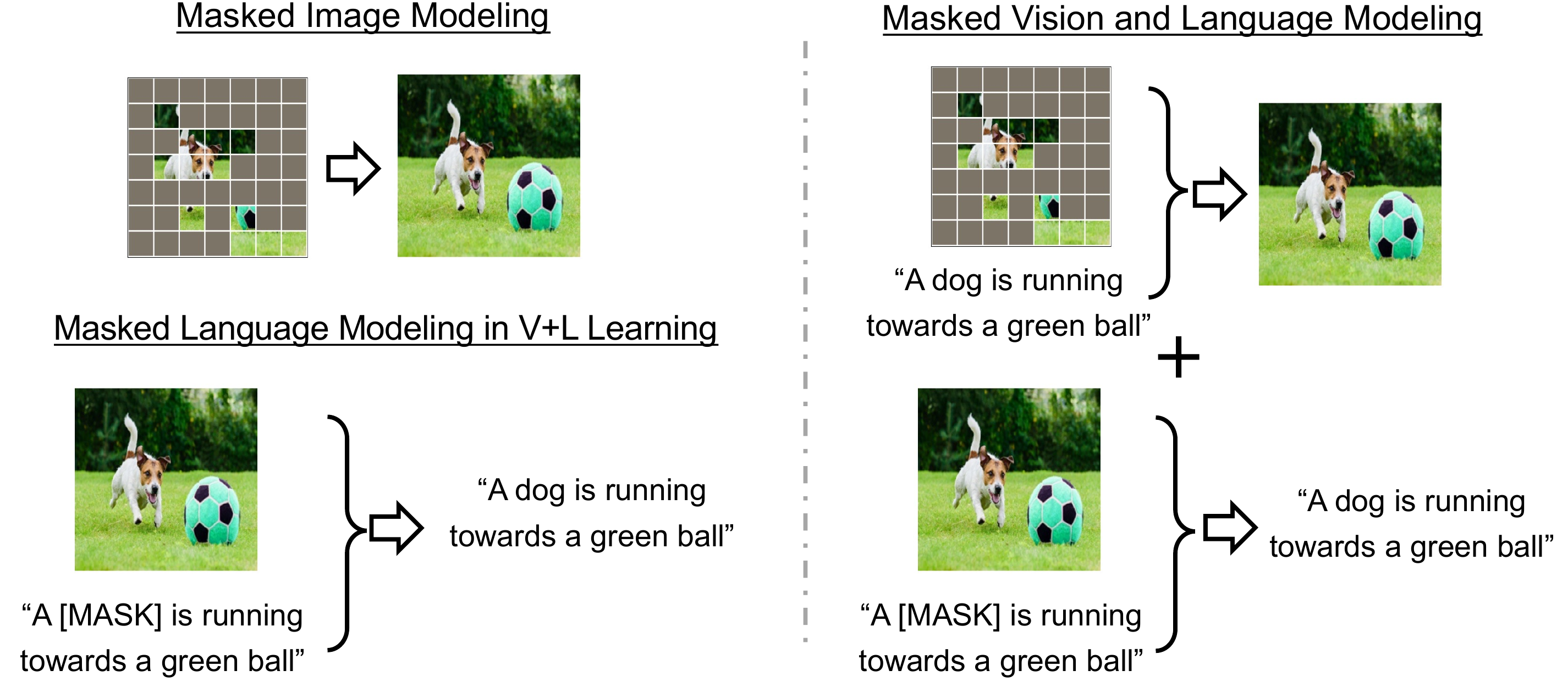}
    \caption{An overview of masked vision and language modeling. The left side shows existing approaches and the right side highlights our proposed approach.}\label{fig:overview}
    \vspace{-0.2cm}
\end{figure}

Masked signal modeling is a popular self-supervisory pre-training task~\citep{devlin2018bert, liu2019roberta, yang2019xlnet, bao2021beit, xie2021simmim, he2022masked}, which aims at reconstructing the masked signals from the unmasked ones. It has been independently explored in the domains of natural language processing (NLP) and computer vision~\citep{devlin2018bert, liu2019roberta, yang2019xlnet, bao2021beit, xie2021simmim, he2022masked}. For example, in the domain of NLP, BERT~\citep{devlin2018bert} and several follow-up works~\citep{liu2019roberta, yang2019xlnet} utilize masked language modeling (MLM) where the model is expected to predict the masked text tokens using unmasked tokens. They have shown that MLM leads to powerful generalization performance across diverse NLP tasks. In computer vision, as shown in Figure~\ref{fig:overview} (top-left), the masked image modeling (MIM) is to predict masked pixels or image patches using unmasked portions of the images. MIM has shown to be an effective pre-training task for learning visual representations~\citep{bao2021beit, xie2021simmim, he2022masked}. 

While MLM and MIM have been actively explored in each domain, existing works do not fully leverage the masked multi-modal signal modeling in the domain of V+L. For example, as shown in Figure~\ref{fig:overview} (bottom-left), several approaches rely only on MLM with unmasked images and do not model the masked images~\citep{duan2022multi, li2022blip, li2021align, li2019visualbert, yang2022vision}. In this case, the distribution of text given image, $p(T | I)$, can be learned, but the distribution of image given text, $P(I | T)$, cannot. This will potentially lead to biased performance in cross-modal retrieval tasks such as image-to-text or text-to-image retrieval as shown in our experiments. Although there exist works that use both modality signals masked, they either use a frozen object detector to extract region-based visual features~\citep{chen2020uniter, li2020unicoder, lu202012, su2019vl, tan2019lxmert} or mask the image tokens from a pre-trained image tokenizer instead of the raw RGB pixels~\citep{dou2022empirical, fu2021violet, singh2022flava}.
These frozen object detector and image tokenizer not only require additional data for training but also prevent the V+L interactions from being learned end-to-end.  

In this paper, we propose joint masked V+L modeling where the original signal is reconstructed by using its masked input and the corresponding unmasked input from the other modality. As illustrated in Figure~\ref{fig:overview} (right part), although we exploit random masking, the dog face in the image can be used to predict the masked text token ``dog'' and the text ``green ball'' can be used to reconstruct the corresponding masked patches in the image. To ensure that the model uses information from both modalities, we explicitly enforce the model to utilize cross-attention to generate the joint representations. Compared with the aforementioned existing works, our approach models both conditional distributions, $p(I|T)$ and $p(T|I)$. Also, the model is trained end-to-end, without frozen bottleneck components that disturb learning interactions between V+L. By reconstructing one modality signal from the corresponding the other modality signal (e.g. reconstructing the text ``dog'' from the visual signals of dog face), the model implicitly learns the alignment between V+L. In addition, we observe that the model trained for the joint masked V+L modeling becomes noticeably effective when the training data is limited. Overall, our contributions are summarized as below:\vspace{-.2cm}
\begin{enumerate}[leftmargin=.7cm]
    \item We propose a joint masked V+L modeling task. We show that models pre-trained with the proposed task, along with common V+L alignment tasks such as image-text matching, achieves state-of-the-art performance on a broad rage of V+L tasks. 
    
    \item We provide a probabilistic interpretation of the proposed method and highlight the difference between ours and existing approaches in terms of the V+L joint distribution estimation.
    
    \item We achieve significantly better performance than other V+L models in the regimes of limited training data, and only $\sim$40\% of data used by the state-of-the-art models is sufficient to match their performance. 
    
\end{enumerate}

\vspace{-.2cm}
\section{Related work}
\begin{figure}[t]
    \centering
    \includegraphics[width=0.85\textwidth]{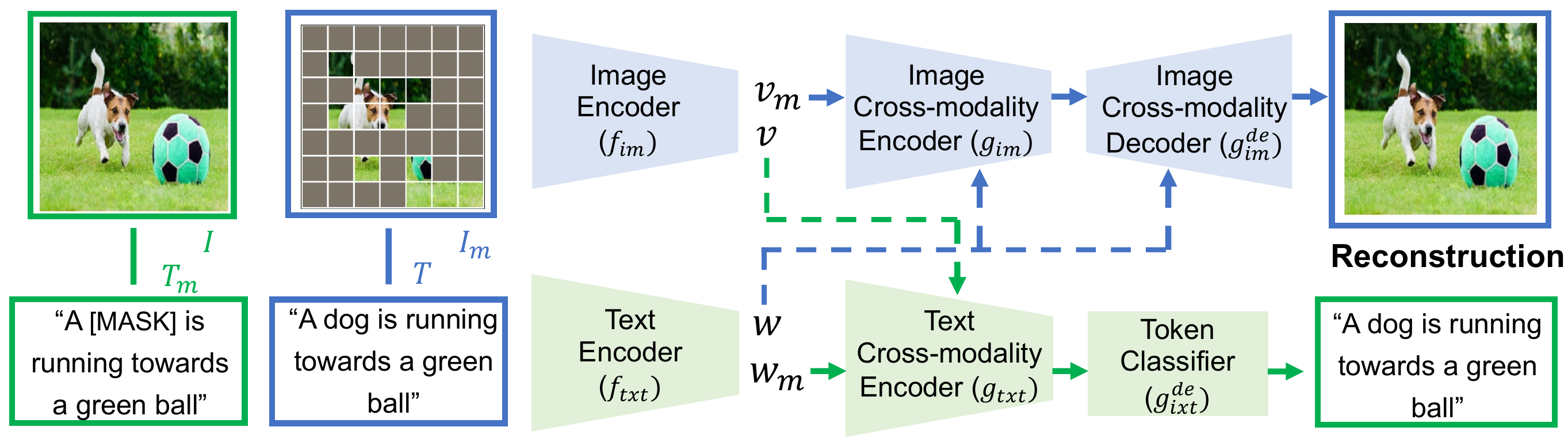}
    \caption{A framework of joint modeling of masked vision and language. The blue and green lines demonstrate the information flow for image and text reconstruction, respectively. The dotted lines indicate the cross-modal input of unmasked signals for generating attention.}\label{fig:framework}\vspace{-0.2cm}
\end{figure}
\begin{figure}[t]
    \centering
    \includegraphics[width=0.6\textwidth]{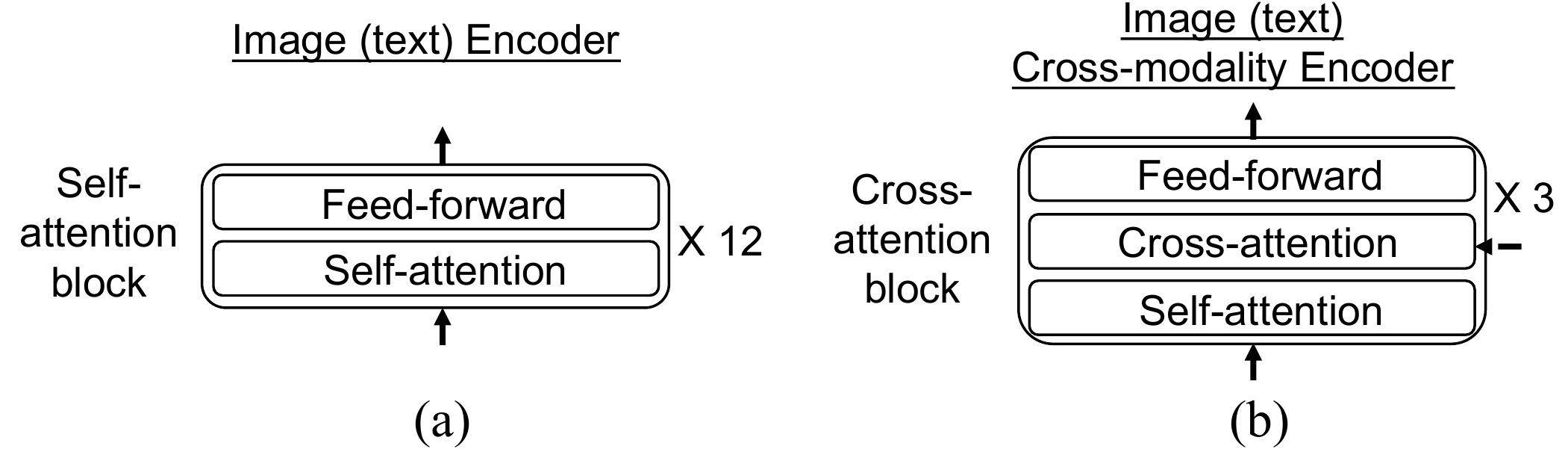}
    \caption{Visualization of image (text) encoders and image (text) cross-modality encoders.}\label{fig:blocks}\vspace{-0.2cm}
\end{figure}
\vspace{-0.2cm}

\paragraph{Vision and language representation learning}
The methods in V+L representation learning can be categorized based on how the  information is fused between the modalities to obtain the joint representations. We group the fusion techniques into three categories: 1) transformers with attention across modalities, 2) contrastive learning with a large-scale pre-training data, 3) a hybrid form of learning with cross-attention and a contrastive loss. The attention across modalities has been widely used with image features extracted from off-the-shelf object detectors and text features obtained from transformer encoders~\citep{chen2020uniter, li2020unicoder, lu202012, tan2019lxmert, zhang2021vinvl, li2020oscar, su2019vl, li2019visualbert}. While cross-attention effectively aligns V+L representations, it is computationally expensive since all possible pairs of images and texts need to be processed. On the contrary, the authors in~\citep{jia2021scaling, radford2021learning, mokady2021clipcap, shen2021much, yuan2021florence} show that contrastive learning with uni-modal encoders and millions of image-text pairs can achieve powerful zero-shot performance in diverse V+L tasks. The contrastive learning-based approaches do not rely on computationally expensive cross-attention but require an excessively large amount of training data. Hence, a combination of contrastive loss and cross-attention is utilized by complementing limitations of both approaches in~\citep{li2021align, li2022blip, yang2022vision, duan2022multi}. In particular, only image and text pairs that result in high similarity by uni-modal encoders are processed using the cross-attention layers to reduce the computational burden and improve the alignment.\vspace{-.3cm}

\paragraph{Masked signal modeling}
is a commonly used pre-training objective in the aforementioned V+L models. It has been independently explored in each of vision and language domain. In the NLP domain, BERT and its variants~\citep{devlin2018bert, liu2019roberta} achieve representations that can generalize to a broad range of NLP tasks through MLM. Autoregressive language models~\citep{radford2018improving, radford2019language} which predict masked future tokens have shown to be effective self-supervised learners. The success of the language models leads to several MIM techniques. BeiT~\citep{bao2021beit} is trained to recover masked visual tokens  which are obtained by a discrete variational autoencoder (dVAE). In SimMIM~\citep{xie2021simmim} and MAE~\citep{he2022masked}, transformers are trained to recover masked patches in an end-to-end fashion. The authors in~\citep{chen2020generative} propose to autoregressively predict the unknown pixels to learn visual representations. In~\citep{bachmann2022multimae}, multiple vision modality data are masked and reconstructed to learn visual representations. In the domain of V+L learning, \citep{arici2021mlim} explores MIM and MLM for catalog data with short text attributes. V+L models with an object detector often aim at recovering only bounding box visual features~\citep{chen2020uniter, li2020unicoder, lu202012, tan2019lxmert, su2019vl}. Several V+L models focus on predicting future text tokens without MIM~\citep{wang2021simvlm, yu2022coca, alayrac2022flamingo}. While both MIM and MLM are explored in \citep{geng2022multimodal}, the trained model is evaluated only on vision tasks. In~\citep{dou2022empirical, fu2021violet, singh2022flava, wang2022image}, image tokens defined by image tokenizers trained with additional 250 million images~\citep{ramesh2021zero} or distillation from the CLIP model~\citep{radford2021learning} trained with 400 million image-text pairs~\citep{peng2022beit} are reconstructed. In our work, we eliminate these model and data dependencies, by directly recovering RGB pixels and text tokens from masked image patches and masked text tokens. Therefore, MIM and MLM are seamlessly integrated to achieve generalizable V+L representations within a simple training framework.

\vspace{-0.3cm}
\section{Method}\vspace{-0.3cm}
Our method has two types of pre-training objectives, which are 1) masked vision and language modeling and 2) multi-modal alignment. We explain each pre-training objective in this section.

\subsection{Masked Vision and Language Modeling}\vspace{-0.2cm}

The overall framework of masked vision and language modeling is shown in Figure~\ref{fig:framework}. We use transformer-based encoders~\citep{vaswani2017attention} for both image and text streams. Given an image-text pair $(I, T)$, an image encoder, $f_{im}$, is used to extract features, $\bm{v} = \{v_{cls}, v_1, ..., v_N\}$, from the image input $I$. $N$ is the number of image patches and $v_{cls}$ is the encoding of the image class token, \texttt{[CLS]}. The text encoder, $f_{txt}$, extracts features, $\bm{w} = \{w_{cls}, w_1, ..., w_M\}$, from the text input, $T$. $M$ is the number of text tokens and $w_{cls}$ is the encoding of the start token of a sentence, \texttt{[START]}. The image and the text encoder consist of 12 self-attention blocks as shown in Figure~\ref{fig:blocks} (a). The image and the text features are further processed by image and text cross-modality encoders. The cross-modality encoders have 3 cross-attention blocks as illustrated in Figure~\ref{fig:blocks} (b). The image (text) cross-modality encoder uses text (image) features to generate attentions. These cross-modality encoders can enhance the representation of one modality by interacting with another modality~\citep{lu202012, tan2019lxmert}.

\textbf{Image and Text Masking:} For text masking, we follow BERT~\citep{devlin2018bert} with minor modifications. In BERT, the original tokens are replaced with either the \texttt{[MASK]} token or random tokens. We use only the \texttt{[MASK]} token to replace tokens to be masked~\citep{wettig2022should}. For image masking, we follow~\citep{he2022masked, xie2021simmim} and use random masking of raw image patches with a masking patch size of $32 \times 32$. Given that $ 224 \times 224$ images are divided into $16 \times 16$ patches for the image encoder, the large masking patch prevents the model from simply copying their neighborhood pixels for reconstruction~\citep{xie2021simmim}.

\textbf{Joint Reconstruction: } We reconstruct the original signals of one modality from its masked input conditioned on the unmasked input of the other modality. Specifically, an original image, $I$, and a masked text, $T_{m}$, are used to reconstruct an original text, $T$, and similarly a masked image, $I_m$, and an original text, $T$, are used to reconstruct the original image, $I$. For image reconstruction, $(I_m, T)$ is first given to the image and the text encoders to obtain masked image features, $\bm{v_m}$, and unmasked text features, $\bm{w}$. Following~\citep{xie2021simmim}, we use both masked and unmasked patches to obtain $\bm{v_m}$. $(\bm{v}_m, \bm{w})$ are further processed by the image cross-modality encoder, $g_{im}$, where $\bm{w}$ is used to compute cross-attentions. The output of $g_{im}$ is mapped back to the original RGB image space by an image cross-modality decoder, $g_{im}^{de}$, which consist of 3 cross-attention blocks followed by a fully connected layer (FC). Although existing work exploits a light-weight transformer decoder with only self-attention~\citep{he2022masked} or a simple linear mapping~\citep{xie2021simmim} for the image decoder, we use joint information between modalities to allow further interactions in decoding. For masked text reconstruction, a token classifier, $g_{txt}^{de}$, which consists of a FC followed by softmax is applied to the output of the text cross-modality encoder, $g_{txt}$, for the token prediction. The masked V+L modeling loss, $\mathcal{L}_{MVLM}$, is defined as\vspace{-.2cm}
\begin{equation}
    \mathcal{L}_{MVLM} =\mathbb{E}_{(I, T) \sim D } [\, \underbrace{\mathcal{H}(y^{M}_T, \phi_{txt}^{M}(I, T_m))}_\text{MLM} +  \underbrace{\dfrac{1}{\Omega({I^M})} \|I^{M} - \phi_{im}^{M}(I_m, T)\|_{1}}_\text{MIM} ]\,, \label{eq:loss_mvlm}
\end{equation}
where $\phi_{txt} = g_{txt}^{de}(g_{txt}(f_{im}(I), f_{txt}(T_{m})))$ and $\phi_{im} = g_{im}^{de}(g_{im}(f_{im}(I_{m}), f_{txt}(T)))$. The loss is computed only for masked pixels and text tokens. Hence, the superscript $M$ denotes data or features correspond to the masked signals. A pair of $I$ and $T$ is sampled from the training dataset $D$, $\mathcal{H}$ denotes cross-entropy, and $y^{M}_T$ is a matrix that contains one-hot row vectors for the ground truth of masked text tokens. $\Omega(\cdot)$ is the number of pixels. When minimizing $\mathcal{L}_{MVLM}$, the model is enforced to reconstruct the original signals by attending to the other modality signals. Cross-attending for reconstruction enables the model to learn interaction between V+L modalities.

\subsection{Multi-modal Alignment}
\label{subsec:alignment}
In addition to the masked signal modeling tasks, we adopt two additional tasks to  explicitly learn multi-modality alignment. The first one is an image-text contrastive (ITC) learning~\citep{radford2021learning, jia2021scaling}. For the $k$-th pair of image and text features out of the image and text encoders, two separate FC layers are used to project the image \texttt{[CLS]} token features and the text \texttt{[START]} token features to the same dimensional feature space with unit norm, $z_{im}^{k}$ and $z_{txt}^{k}$, respectively. The loss, $\mathcal{L}_{ITC}$ is computed as
\vspace{-.1cm}
\begin{equation}
    \mathcal{L}_{ITC} = - \dfrac{1}{N}\sum_{k=1}^{N}\left[\log \dfrac{\exp (z_{im}^{k} \cdot z_{txt}^{k} / \tau)}{\sum_{n=1}^{N} \exp (z_{im}^{k} \cdot z_{txt}^{n} / \tau)} + \log \dfrac{\exp (z_{im}^{k} \cdot z_{txt}^{k} / \tau)}{\sum_{n=1}^{N} \exp (z_{im}^{n} \cdot z_{txt}^{k} / \tau)} \right],
\end{equation}

where $N$ and $\tau$ are the batch size and the temperature scaling parameter, respectively. The second task is an image-text matching (ITM)~\citep{chen2020uniter, li2021align, li2020oscar}, predicting whether an image and a text are aligned or not. The \texttt{[CLS]} and \texttt{[START]} token features from the image and text cross-modality encoders are $z_{im}^{cross}$ and $z_{txt}^{cross}$, respectively. To fuse these two features, we compute the element-wise product of $z_{im}^{cross}$ and $z_{txt}^{cross}$~($z_{im}^{cross} * z_{txt}^{cross}$), and a FC layer followed by softmax is applied to obtain the final prediction~\citep{lu2019vilbert}. For training, we use $y_{ITM} =1$,  when $z_{im}^{cross}$ and $z_{txt}^{cross}$ are a pair. Otherwise, $y_{ITM} = 0$. The loss, $\mathcal{L}_{ITM}$, is defined as
\begin{equation}
    \mathcal{L}_{ITM} = \mathbb{E}_{(I, T) \sim D } [ \mathcal{H}(y_{ITM}, g^{itm}(z_{im}^{cross} * z_{txt}^{cross}))].
\end{equation}
Following~\citep{li2021align}, we sample in-batch hard negatives based on the distribution of the cosine similarity between $z_{im}$ and $z_{txt}$. The overall pre-training loss, $\mathcal{L}$, is defined as $\mathcal{L} = \mathcal{L}_{MVLM} + \mathcal{L}_{ITC} + \mathcal{L}_{ITM}.$ We term our model trained with loss $\mathcal{L}$ as MaskVLM (\textbf{Mask}ed \textbf{V}ision and \textbf{L}anguage \textbf{M}odeling).

\vspace{-.1cm}
\subsection{Probabilistic interpretation}\vspace{-.1cm}
We differentiate MaskVLM from the existing V+L models using masked signal modeling from a perspective of likelihood estimation. The training objective of masked signal modeling on uni-modal signals, $X$, focuses on learning the data distribution $p(X)$ which is formulated by the law of total probability as 
$p(X) = \sum_{X_{m} \in \mathcal{M_{X}}} p(X_{m}) \cdot p(X | X_{m})$,
where $X_{m}$ is an instance of  masked signal from the set of all possible masked signals, $\mathcal{M}_{X}$. MIM or MLM learns the data distribution by maximizing $\sum_{X_{m} \in \mathcal{M_{X}}} p(X | X_{m})$~\citep{bengio2013generalized}. 

In V+L representation learning, the ultimate goal is to learn the joint distribution of multi-modal signals, $p(I, T)$. However, the authors in~\citep{sohn2014improved} pointed out that directly maximizing the likelihood for the joint distribution is challenging because of the heterogeneous multi-modal data distributions. Instead, they show minimizing variation of information defined as $- \mathbb{E}_{(I, T) \sim D }(\log p(I |T) + \log p(T |I))$ is sufficient to estimate the joint distribution. From a perspective of variation of information, the limitations in existing works can be better understood. Several existing works attempted to approximate the joint distribution using MLM with unmasked image~\citep{duan2022multi, li2021align, li2019visualbert, yang2022vision}. In other words,  $p(T | I, T_{m})$ is maximized to learn the conditional distribution, $p(T | I)$, but $p(I | T)$ is not modeled. In other existing works~\citep{chen2020uniter, li2020unicoder, lu202012, su2019vl, tan2019lxmert}, where both modalities are masked, the visual masking is limited to mask the visual features extracted from a frozen object detector, $\psi(\cdot)$, instead of the raw image pixels. In this case, the distributions $p(\psi(I) | T)$ and $p(T | \psi(I))$ are modeled instead of $p(I | T)$ and $p(T | I)$. This frozen feature extractor can bottleneck the direct estimation of the underlying data distribution. MaskVLM is trained end-to-end to estimate both conditional distributions, $p(I | T)$ and $p(T | I)$, which directly minimizes the variation of information. We hypothesize this modeling of conditional distributions for both modalities could lead to superior performance in both large-scale and limited data training scenarios, which we empirically demonstrated in Section~\ref{sec:experiments}.\vspace{-0.2cm}

\section{Experiments}\label{sec:experiments}\vspace{-.1cm}
\subsection{Pre-training datasets and downstream tasks}\label{subsec:datasets}

We use the union of four datasets for pre-training so that we can perform  a fair comparison with existing state-of-the-art methods~\citep{chen2020uniter, li2021align}. These datasets are  Conceptual Captions (CC)~\citep{sharma2018conceptual}, SBU Captions~\citep{ordonez2011im2text}, Visual Genome (VG)~\citep{krishna2017visual}, and COCO Captions~\citep{lin2014microsoft}. While VG and COCO contain captions annotated by humans, CC and SBU Captions are automatically collected from the web. The total number of unique images and image-text pairs in the four datasets are 4.1M and 5.2M, respectively. We term this pre-training dataset as the 4M dataset.  We validate the pre-trained model on the following four downstream tasks:

\textbf{Image-Text Retrieval: } We perform text-to-image and image-to-text retrieval.  We use the ITC and ITM losses of Section \ref{subsec:alignment} for finetuning and the finetuned models are evaluated on COCO~\citep{lin2014microsoft} and Flickr30k~\citep{plummer2015flickr30k}. In addition, since COCO is used for pre-training, zero-shot retrieval performance is reported on Flickr30k. In~\citep{li2021align}, the model finetuned on COCO is used for the zero-shot evaluation on Flickr30k. Although it may result in better performance, we believe that using the finetuned model does not validate the zero-shot capability of the pre-trained model. Therefore, we use the pre-trained model directly for zero-shot evaluation. Following~\citep{li2021align}, we first retrieve top-$k$ candidates using the similarity scores from the image and the text encoders. The top-$k$ candidates are further processed by the cross-modality encoders to obtain the final retrieval results.

\textbf{Visual Question Answering (VQA): } Here, given an image and a question pair, the model should generate a correct answer. The model is evaluated on VQA v2~\citep{goyal2017making}. We adopt the answer generation framework~\citep{cho2021unifying} and finetune the base model with a fusion encoder and an answer decoder. The model architectures are visualized in Figure~\ref{fig:vqa_nlvr} (a) of Appendix. The fusion encoder consists of one cross-attention block shown in Figure~\ref{fig:blocks} (b). The output from the text cross-modality encoder is used as queries, and the image cross-modality encoder output is utilized to create attentions in the fusion encoder. The architecture of the answer decoder is the same as that of the text cross-modality encoder, but it is trained with a language modeling loss to generate the answers. Specifically, the output of the fusion encoder is used for computing attentions and the answer tokens are autoregressively predicted. During inference, \texttt{[START]} token is used as an initial token to generate following answer tokens. 

\textbf{Natural Language for Visual Reasoning (NLVR):} This tasks involves a binary classification with a triplet, (text, image1, image2). The goal here is to predict whether the text describes the pair of images. For finetuning, we feedforward (text, image1) and (text, image2) separately to extract the features as shown in Figure~\ref{fig:vqa_nlvr} (b). The \texttt{[CLS]} token features of image1 and image2 from the image encoder are denoted as $v_1$ and $v_2$, respectively. The \texttt{[START]} token text features from the text encoder is $w$. These features are processed by the cross-modality encoders. The outputs of the image and text cross-modality encoders are fused by element-wise multiplication. The fused features for both images are concatenated, and a classifier with two linear layers predicts whether the text is aligned with the image pair or not. NLVR2~\citep{suhr2018corpus} is used for the evaluation.

\textbf{Visual Entailment (VE):} Given an image text pair, the task is to classify the relationship between the image and the text into one of three categories: entailment, neutral, and contradictory. The element-wise product of the output from the image and the text cross-modality encoders is forwarded to a classifier of two linear layers for prediction. SNLI-VE~\citep{xie2019visual} is used for evaluation.\vspace{-0.2cm}

\begin{table}[t]
\aboverulesep = 0.48mm
\belowrulesep = 0.48mm
\small
\centering
\setlength\tabcolsep{4pt}
\resizebox{\textwidth}{!}{%
\begin{tabular}{cccccccc|cccccc}
\toprule
\multirow{3}{*}{Method} & \multirow{3}{*}{\# images} & \multicolumn{6}{c}{MSCOCO (5K)} & \multicolumn{6}{c}{Flickr30k (1K)} \\
 &  & \multicolumn{3}{c}{Image Retrieval} & \multicolumn{3}{c}{Text Retrieval} & \multicolumn{3}{c}{Image Retrieval} & \multicolumn{3}{c}{Text Retrieval} \\
 &  & R@1 & R@5 & R@10 & R@1 & R@5 & R@10 & R@1 & R@5 & R@10 & R@1 & R@5 & R@10 \\ \midrule
ImageBERT~\citep{qi2020imagebert} & 6M & 50.5 & 78.7 & 87.1 & 66.4 & 89.8 & 94.4 & 73.1 & 92.6 & 96.0 & 87.0 & 97.6 & 99.2 \\
UNITER~\citep{chen2020uniter} & 4M & 52.9 & 79.9 & 88.0 & 65.7 & 88.6 & 93.8 & 75.6 & 94.1 & 96.8 & 87.3 & 98.0 & 99.8 \\
VILLA~\citep{gan2020large} & 4M & - & - & - & - & - & - & 76.3 & 94.2 & 96.8 & 87.9 & 97.5 & 98.8 \\
OSCAR~\citep{li2020oscar} & 4M & 54.0 & 80.8 & 88.5 & 70.0 & 91.1 & 95.5 & - & - & - & - & - & - \\ 
ALBEF~\citep{li2021align} & 4M & 56.8 & 81.5 & 89.2 & 73.1 & 91.4 & 96.0 & 82.8 & \textbf{96.7} & \textbf{98.4} & 94.3 & \textbf{99.4} & 99.8 \\ 
Triple~\citep{yang2022vision} & 4M & 59.0 & 83.2 & 89.9 & 75.6 & 92.8 & 96.7 & 84.0 & \textbf{96.7} & \textbf{98.5} & 94.9 & \textbf{99.5} & 99.8 \\ 
Codebook~\citep{duan2022multi} & 4M & 58.7 & 82.8 & 89.7 & 75.3 & 92.6 & 96.6 & 83.3 & 96.1 & 97.8 & 95.1 & \textbf{99.4} & \textbf{99.9} \\ 
\rowcolor{lightgray} ALIGN~\citep{jia2021scaling} & 1.8B & 59.9 & 83.3 & 89.8 & 77.0 & 93.5 & 96.9 & 84.9 & 97.4 & 98.6 & 95.3 & 99.8 & 100.0 \\ \midrule
MaskVLM & 4M & \textbf{60.1} & \textbf{83.6} & \textbf{90.4} & \textbf{76.3} & \textbf{93.8} & \textbf{96.8} & \textbf{84.5} & \textbf{96.7} & \textbf{98.2} & \textbf{95.6} & \textbf{99.4} & \textbf{99.9} \\ \bottomrule
\end{tabular}}\vspace{0.3cm}\caption{Comparison with finetuned state-of-the-art methods on image-text retrieval. The gray row indicates that the model is trained with significantly larger number of data than MaskVLM.}\label{tab:finetune_retrieval}
\end{table}

\begin{table}[t]
\centering
\small
\centering
\setlength\tabcolsep{4pt}
\resizebox{0.7\textwidth}{!}{%
\begin{tabular}{cccccccc}
\toprule
\multirow{3}{*}{Method} & \multirow{3}{*}{\# images} & \multicolumn{6}{c}{Flickr30k (1K)} \\
 &  & \multicolumn{3}{c}{Image Retrieval} & \multicolumn{3}{c}{Text Retrieval} \\
 &  & R@1 & R@5 & R@10 & R@1 & R@5 & R@10 \\ \midrule
ImageBERT~\citep{qi2020imagebert} & 6M & 54.3 & 79.6 & 87.5 & 70.7 & 90.2 & 94.0 \\
Unicoder-VL~\citep{li2020unicoder} & 3.8M & 48.4 & 76.0 & 85.2 & 64.3 & 85.8 & 92.3 \\
ViLT~\citep{kim2021vilt} & 4M & 55.0 & 82.5 & 89.8 & 73.2 & 93.6 & 96.5 \\
UNITER~\citep{chen2020uniter} & 4M & 66.2 & 88.4 & 92.9 & 80.7 & 95.7 & 98.0 \\
ALBEF~\citep{li2021align} & 4M & 68.2 & 88.6 & 93.0 & 84.9 & 97.2 & 99.0 \\
FLAVA~\citep{singh2022flava} & 68M & 65.2 & 89.4 & - & 67.7 & 94.0 & - \\
\rowcolor{lightgray}CLIP~\citep{radford2021learning} & 400M & 68.7 & 90.6 & 95.2 & 88.0 & 98.7 & 99.4 \\
\rowcolor{lightgray}ALIGN~\citep{jia2021scaling} & 1.8B & 75.7 & 93.8 & 96.8 & 88.6 & 98.7 & 99.7 \\ \midrule
MaskVLM & 4M & \textbf{75.0} & \textbf{92.5} & \textbf{95.8} & \textbf{87.0} & \textbf{97.9} & \textbf{99.3} \\ 
 \bottomrule
\end{tabular}}\vspace{0.3cm}\caption{Zero-shot image-text retrieval performance on Flickr30k. The gray row indicates that the model is trained with significantly larger number of data than MaskVLM.}\label{tab:zeroshot}
\end{table}

\subsection{Implementation details}\label{subsec:implementaton_detials}
We use a Visual Transformer (ViT)~\citep{dosovitskiy2020image} pre-trained on ImageNet~\citep{deng2009imagenet} and a pre-trained RoBERTa from~\citep{liu2019roberta} to initialize the image and the text encoder, respectively. We pre-train the model for 50 epochs when the 4M dataset is used and 30 epochs for all other experiments. A batch size of $512$ is used with $16$ NVIDIA Tesla V100 GPUs. All parameters are optimized using AdamW ~\citep{loshchilov2017decoupled} with a weight decay of $0.05$. Following~\citep{xie2021simmim}, we use the image masking ratio of 60\%. While 15\% masking ratio is used for text in language models~\citep{devlin2018bert, liu2019roberta}, we use 30\% since the paired image can provide additional information for text reconstruction. During pre-training, the learning rate is warmed up to $3 \times 10^{-4}$ in the first $5$ epochs and decayed to $3 \times 10^{-5}$ using a cosine scheduler. The learning rates for the image encoder and the text encoder are set to $10^{-5}$, which is less than that of the cross-modality encoders. An image size of $224 \times 224$ and RandAugment~\citep{cubuk2020randaugment} are used. During finetuning, the image  is resized to $384 \times 384$ and the positional encoding is interpolated following~\citep{dosovitskiy2020image}. More details can be found in Appendix.

\begin{table}[t]
\small
\centering
\setlength\tabcolsep{4pt}
\resizebox{0.73\textwidth}{!}{%
\begin{tabular}{cccccccc}
\toprule
\multirow{2}{*}{Method} & \multirow{2}{*}{\# images} & \multicolumn{2}{c}{VQA} & \multicolumn{2}{c}{NLVR2} & \multicolumn{2}{c}{SNLI-VE} \\
& & test-dev & test-std & dev & test-P & val & test \\ \midrule
VisualBERT~\citep{li2019visualbert} & 113K & 70.80 & 71.00 & 67.40 & 67.00 & - & - \\
VL-BERT~\citep{su2019vl} & 3.3M & 71.16 & - & - & - & - & - \\
LXMERT~\citep{tan2019lxmert} & 180K & 72.42 & 72.54 & 74.90 & 74.50 & - & - \\
12-in-1~\citep{lu202012} & 3.3M & 73.15 & - & - & 78.87 & - & 76.95 \\
UNITER~\citep{chen2020uniter} & 4M & 72.70 & 72.91 & 77.18 & 77.85 & 78.59 & 78.28 \\
VL-BART/T5~\citep{cho2021unifying} & 180K & - & 71.30 & - & 73.60 & - & - \\
ViLT~\citep{kim2021vilt} & 4M & 70.94 & - & 75.24 & 76.21 & - & - \\
OSCAR~\citep{li2020oscar} & 4M & 73.16 & 73.44 & 78.07 & 78.36 & - & - \\
VILLA~\citep{gan2020large} & 4M & 73.59 & 73.67 & 78.39 & 79.30 & 79.47 & 79.03 \\
ALBEF~\citep{li2021align} & 4M & 74.54 & 74.70 & 80.24 & 80.50 & 80.14 & 80.30 \\
Triple~\citep{yang2022vision} & 4M & 74.90 & 74.92 & 80.54 & 80.11 & \textbf{80.51} & 80.29 \\
Codebook~\citep{duan2022multi} & 4M & 74.86 & 74.97 & 80.50 & 80.84 & \textbf{80.47} & 80.40 \\ 
FLAVA~\citep{singh2022flava} & 68M & 72.80 & - & - & - & - & 79.00 \\
\rowcolor{lightgray}$\text{SimVLM}_{base}$ \citep{wang2021simvlm} & 1.8B & 77.87 & 78.14 & 81.72 & 81.77 & 84.20 & 84.15 \\
\midrule
MaskVLM & 4M & \textbf{75.45} & \textbf{75.40} & \textbf{81.58} & \textbf{81.98} & \textbf{80.37} & \textbf{80.67} \\ \bottomrule
\end{tabular}}\vspace{0cm}\caption{Comparison with state-of-the-art methods on VQA, NLVR2, and VE. The gray row indicates that the model is trained with significantly larger number of data than MaskVLM.}\label{tab:vqa_nlvr_ve}
\end{table}
\subsection{Evaluation on Image-Text Retrieval, VQA, NLVR, and VE}\vspace{-.2cm}

We compare the finetuned image-text retrieval performance of the proposed MaskVLM with the state-of-the-art methods in Table~\ref{tab:finetune_retrieval}. The second column is the number of unique images used for pre-training and the retrieval performance is evaluated in terms of Recall@k (R@k). We do not directly compare with ALIGN~\citep{jia2021scaling} since it is trained with more than 300 times of data used for MaskVLM. However, we still highlight the small performance gap between MaskVLM and ALIGN. We achieve the best performance in all Recall@k metrics on both COCO and Flickr30k except for the image retrieval R@10 and text retrieval R@5 on Flickr30k. Compared to ALIGN, we even achieve higher R@1 for image retrieval on COCO and text retrieval on Flickr30k. Table~\ref{tab:zeroshot} shows the zero-shot retrieval performance of the state-of-the-art methods on Flickr30k. MaskVLM achieves a significant improvement over the second best method, ALBEF~\citep{li2021align}, by 6.8 points at R@1 for image retrieval. Given that ALBEF is not trained for MIM, we hypothesize that ALBEF achieves the biased performance for text retrieval and MaskVLM achieves the significant improvement in image retrieval by additional MIM which models $p(I | T)$. While FLAVA exploits both MLM and MIM with the pre-trained image tokenizer, using 13 times more data than MaskVLM, MaskVLM still outperforms FLAVA by 9.8 and 19.3 points at R@1 for image and text retrieval respectively. Compared with CLIP \citep{radford2021learning} which is trained with at least 76 times more data than MaskVLM, we still achieve higher R@1 for image retrieval by 6.3 points. In general, MaskVLM achieves state-of-the-art performance in both finetuning and zero-shot experiments.

We report the accuracies on VQA, NLVR, and VE in Table~\ref{tab:vqa_nlvr_ve}. Except for SimVLM whose pre-training data is more than 300 times larger than that of MaskVLM, we consistently achieve the best performances in all these tasks except for the validation split of NLVR2. In particular, MaskVLM is better than the second best method by 0.43, 1.14, and 0.27 on the test splits of VQA, NLVR2, and SNLI-VE, respectively. Compared to the base model of SimVLM, we narrow the accuracy gaps to $2.74\%$ and $3.48\%$ in test-std and test splits of VQA and VE, respectively. MaskVLM achieves higher accuracy than $\text{SimVLM}_{base}$ in the test split of NLVR2 by $0.21\%$.

\begin{table}[t]
\small
\centering
\setlength\tabcolsep{4pt}
\resizebox{0.8\textwidth}{!}{%
\begin{tabular}{ccccccccccc}
\toprule
\multirow{2}{*}{\begin{tabular}[c]{@{}c@{}}Dataset (\# of samples)\end{tabular}} & \multirow{2}{*}{Method} & \multicolumn{2}{c}{COCO IR} & \multicolumn{2}{c}{COCO TR} & VQA & \multicolumn{2}{c}{NLVR2} & \multicolumn{2}{c}{SNLI-VE} \\
 &  & \multicolumn{1}{c}{R@1} & R@5 & \multicolumn{1}{c}{R@1} & R@5 & dev & \multicolumn{1}{c}{dev} & test-P & \multicolumn{1}{c}{val} & test \\ \midrule
\multirow{2}{*}{\begin{tabular}[c]{@{}c@{}}CC 50\% + COCO (2M)\end{tabular}} & ALBEF & \multicolumn{1}{c}{49.97} & 77.35 & \multicolumn{1}{c}{65.76} & 89.32 & 73.07 & \multicolumn{1}{c}{76.58} & 76.89 & \multicolumn{1}{c}{79.20} & 79.19 \\ 
 & Ours & \multicolumn{1}{c}{\textbf{56.36}} & \textbf{81.98} & \multicolumn{1}{c}{\textbf{73.22}} & \textbf{92.00} & \textbf{74.24} & \multicolumn{1}{c}{\textbf{79.81}} & \textbf{79.47} & \multicolumn{1}{c}{\textbf{79.69}} & \textbf{79.50} \\ \midrule
\multirow{2}{*}{\begin{tabular}[c]{@{}c@{}}CC 25\% + COCO (1.3M)\end{tabular}} & ALBEF & \multicolumn{1}{c}{48.09} & 75.63 & \multicolumn{1}{c}{64.24} & 87.20 & 72.65 & \multicolumn{1}{c}{75.20} & 76.78 & \multicolumn{1}{c}{78.84} & 78.96 \\
 & Ours & \multicolumn{1}{c}{\textbf{55.11}} & \textbf{81.28} & \multicolumn{1}{c}{\textbf{72.18}} & \textbf{91.48} & \textbf{74.17} & \multicolumn{1}{c}{\textbf{79.13}} & \textbf{78.94} & \multicolumn{1}{c}{\textbf{79.35}} & \textbf{79.76} \\ \midrule
\multirow{2}{*}{\begin{tabular}[c]{@{}c@{}}CC 10\% + COCO (0.9M)\end{tabular}} & ALBEF & \multicolumn{1}{c}{45.33} & 73.57 & \multicolumn{1}{c}{61.00} & 84.98 & 72.14 & \multicolumn{1}{c}{74.81} & 74.64 & \multicolumn{1}{c}{78.51} & 78.36 \\ 
 & Ours & \multicolumn{1}{c}{\textbf{54.04}} & \textbf{80.74} & \multicolumn{1}{c}{\textbf{70.04}} & \textbf{91.24} & \textbf{73.93} & \multicolumn{1}{c}{\textbf{78.62}} & \textbf{77.19} & \multicolumn{1}{c}{\textbf{79.11}} & \textbf{79.60} \\ \bottomrule
\end{tabular}\caption{Downstream task performance with limited pre-training data.}\label{tab:limited}}
\end{table}
\vspace{-.2cm}
\subsection{Evaluation with limited pre-training data}\label{subsec:limited}\vspace{-.2cm}
\begin{wrapfigure}{r}{0.48\textwidth}
\centering
\begin{minipage}[t]{0.495\linewidth}
    \centering
    \hspace{-.5cm}
    \includegraphics[width=\linewidth]{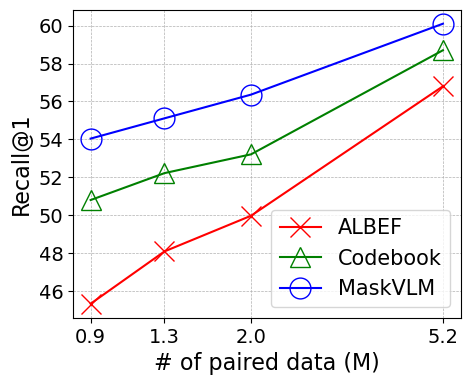}
\end{minipage}\hspace{-0.2cm}
\begin{minipage}[t]{0.495\linewidth}
    \centering
    \hspace{-0.4cm}
    \includegraphics[width=\linewidth]{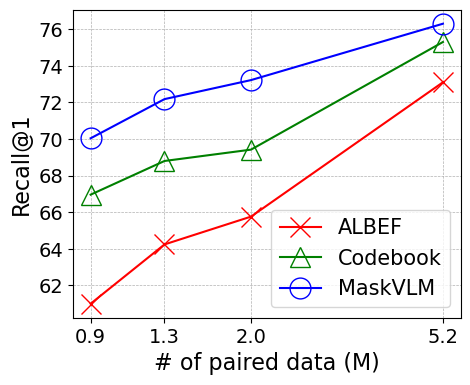}
\end{minipage}\caption{R@1 plots for image retrieval (left) and text retrieval (right) on COCO using limited pre-training data.}\label{fig:recall}
\end{wrapfigure}We highlight the performance of MaskVLM in limited data scenarios. In particular, we create three subsets of the 4M pre-training data by sampling 50\%, 25\%, and 10\% of CC and combining them with COCO. The number of image-text pairs in each subset is around 39\%, 25\%, and 16\% of the 4M pre-training data which contain 5.2M pairs, respectively. We pre-train models with these subsets of the data and analyze the downstream task performance in comparison with state-of-the-art methods. The results are reported in Table~\ref{tab:limited}. Particularly, image and text retrieval R@1 performance on COCO is also visualized in Figure~\ref{fig:recall}. We compare MaskVLM with the most recent state-of-the-art methods, which are ALBEF~\citep{li2021align} and Codebook~\citep{duan2022multi}. In Table~\ref{tab:limited}, as the size of pre-training data becomes smaller from CC 50\% + COCO to CC 10\% + COCO, the performance gap between MaskVLM and ALBEF increases from 6.39 to 8.71 at R@1 in COCO image retrieval (IR), 7.46 to 9.04 at R@1 in COCO text retrieval (TR), 1.17 to 1.79 in VQA and 0.31 to 1.24 in the test set of SNLI-VE. In NLVR2 and VQA, MaskVLM trained with CC 10\% + COCO achieves higher accuracy than ALBEF trained with CC50\% + COCO, which contains more than twice of image-text pairs in CC 10\% + COCO. In Figure~\ref{fig:recall}, while Codebook shows competitive recall performance compared to the MaskVLM with the 4M dataset (~5.2M pairs), the R@1 differences in image and text retrieval, respectively, increase from 1.4 and 1.0 in the 4M dataset to 3.15 and 3.80 in CC 50\% + COCO. Our model trained with CC25\% + COCO  outperforms Codebook trained with CC50\% + COCO by 1.90 and 2.76 points in terms of image and text retrieval R@1, respectively. Since one of the main differences in MaskVLM compared to ALBEF and Codebook is the additional MIM, we believe that joint modeling of V+L contribute to better performance in limited data scenarios. 

\subsection{Ablation study}\vspace{-0.2cm}
We perform an ablation study using different combinations of loss components to highlight the contribution of masked V+L modeling. We compare six models with the same architecture but with different loss functions for pre-training. We pre-train all models on the CC 50\% + COCO dataset and compare finetuned and zero-shot retrieval performance on Flickr30k in Table~\ref{tab:ablation_loss}. We note that zero-shot evaluation of the MLM + MIM model cannot be performed because the FC layers to compute ITM and ITC are not trained during pre-training. ITC and ITM are closely related to the retrieval task since they are used for finetuning and measuring the similarity between images and texts. However, MLM + MIM still achieves significantly better finetuned and zero-shot performance than ITC, which shows that MLM + MIM alone learns meaningful V+L representations. Compared to ITC+ ITM in the finetuned performance, ITC + ITM + MLM achieves slightly improved R@1 by 0.38 in image retrieval and degraded R@1 by 0.3 in text retrieval. When MIM alone is used with ITC + ITM as well, finetuned R@1 is improved by 0.16 and degraded by 0.8 for image and text retrieval, respectively, over ITC + ITM. On the other hand, when ITC + ITM + MLM + MIM is used, the model achieves significant improvement of finetuned performance over ITC + ITM + MLM by 0.92 and 2.10 for R@1 image and text retrieval, respectively. ITC + ITM + MLM + MIM also obtains the best performance in zero-shot retrieval. This result further supports the advantage of joint modeling for masked V+L signals. 
\begin{table}[t]
\small
\centering
\setlength\tabcolsep{4pt}
\resizebox{0.75\textwidth}{!}{%
\begin{tabular}{ccccccccc}
\toprule
\multirow{3}{*}{Loss} & \multicolumn{4}{c}{Finetuned} & \multicolumn{4}{c}{Zero-shot} \\
 & \multicolumn{2}{c}{IR} & \multicolumn{2}{c}{TR} & \multicolumn{2}{c}{IR} & \multicolumn{2}{c}{TR} \\
 & R@1 & R@5 & R@1 & R@5 & R@1 & R@5 & R@1 & R@5 \\ \midrule
ITC & 65.10 & 89.88 & 80.10 & 96.90 & 55.08 & 80.90 & 68.40 & 90.00 \\
ITC + ITM & 79.96 & 95.56 & 92.30 & 98.90 & 69.50 & 89.54 & 82.40 & 96.60 \\
MLM +MIM & 76.08 & 94.40 & 90.30 & 98.80 & - & - & - & - \\
ITC + ITM + MLM & 80.34 & 95.82 & 92.00 & 99.30 & 70.74 & 90.92 & 84.40 & 97.10 \\
ITC + ITM + MIM & 80.12 & 95.56 & 91.50 & 99.00 & 69.26 & 90.30 & 82.90 & 97.20 \\
ITC + ITM + MLM + MIM & \textbf{81.26} & \textbf{96.00} & \textbf{94.10} & \textbf{99.60} & \textbf{71.18} & \textbf{91.12} & \textbf{85.60} & \textbf{97.50} \\ \bottomrule
\end{tabular}}\caption{Image-text retrieval evaluation on Flickr30k with different loss functions for pre-training.}\label{tab:ablation_loss}
\end{table}

\vspace{-0.2cm}
\subsection{Qualitative results}
\begin{figure}[t]
    \centering
    \includegraphics[width=0.95\textwidth]{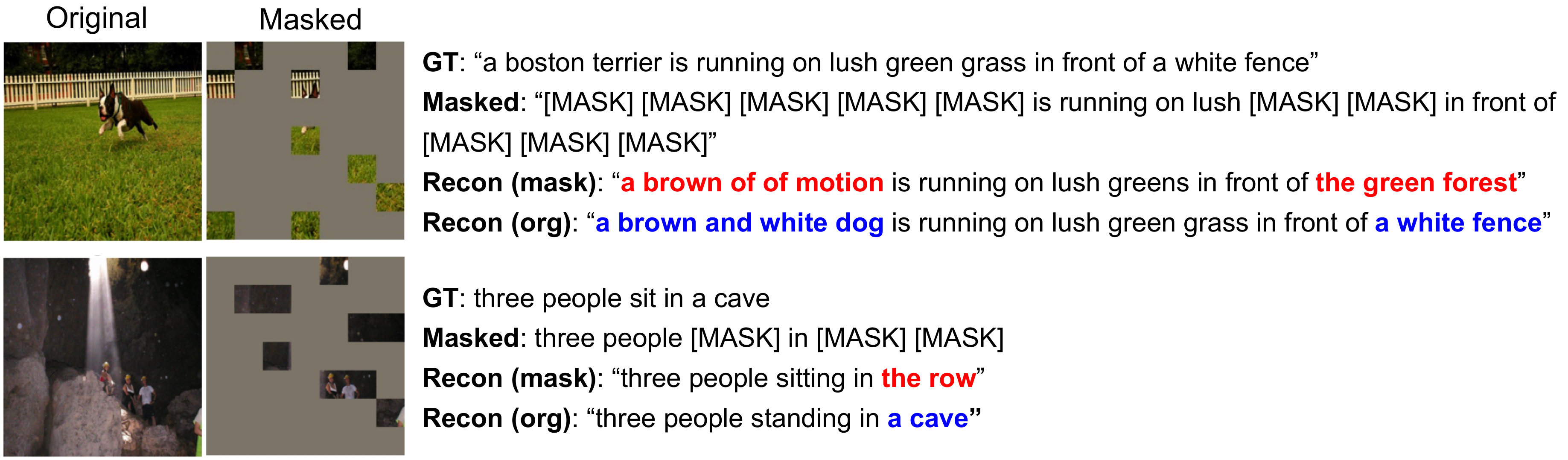}
    \caption{Masked language modeling examples using masked and original images. ``Recon (mask)'' and ``Recon (org)'' denote reconstructed text from the masked image and the original image, respectively.}\label{fig:example}
\end{figure}

We perform a qualitative analysis to show the role of multi-modal information in the reconstruction of masked signals from our model. To be specific, we illustrate the prediction of masked text tokens with and without the corresponding images. This illustration highlights how MaskVLM effectively utilizes both modality information to complete the masked signal modeling task. Figure~\ref{fig:example} shows the reconstruction of masked texts using original images (``Recon (org)'') and masked images (``Recon (mask)''). In the first top example, when the model is given a masked text and a masked image which does not contain the ``dog'', the reconstruction is performed by using only available information such as image patches of ``green grass''. Thus, the prediction is limited to ``a brown of motion'' or ``the green forest''. However, when the original image is used for reconstruction, both ``a brown and white dog'' and ``white fence'' are reconstructed by accurately attending to the image. In the bottom example, the visible patches of the masked image contain a few people, but lack background information. Consequently, the reconstruction with the masked image does not contain any background information but the background ``cave'' is reflected in the reconstruction with the original image. Theses examples confirm that MaskVLM has learned to perform masked signal modeling using both V+L information.

\vspace{-0.2cm}
\section{Conclusion}\vspace{-0.2cm}
We propose masked vision and language modeling as a pre-training task for learning V+L representations. We provide a probabilistic interpretation to highlight the contribution of the proposed method and validate its advantages in both large-scale and limited data regimes. We consistently achieve the state-of-the-art performance in a broad range of V+L tasks.

\newpage
\subsubsection*{Acknowledgments}
We thank Jiali Duan for providing results of the Codebook~\citep{duan2022multi} with limited pre-training data in Figure~\ref{fig:recall}.

\bibliography{iclr2023_conference}

\begin{thebibliography}{73}
\providecommand{\natexlab}[1]{#1}
\providecommand{\url}[1]{\texttt{#1}}
\expandafter\ifx\csname urlstyle\endcsname\relax
  \providecommand{\doi}[1]{doi: #1}\else
  \providecommand{\doi}{doi: \begingroup \urlstyle{rm}\Url}\fi

\bibitem[Alayrac et~al.(2022)Alayrac, Donahue, Luc, Miech, Barr, Hasson, Lenc,
  Mensch, Millican, Reynolds, et~al.]{alayrac2022flamingo}
Jean-Baptiste Alayrac, Jeff Donahue, Pauline Luc, Antoine Miech, Iain Barr,
  Yana Hasson, Karel Lenc, Arthur Mensch, Katie Millican, Malcolm Reynolds,
  et~al.
\newblock Flamingo: a visual language model for few-shot learning.
\newblock \emph{arXiv preprint arXiv:2204.14198}, 2022.

\bibitem[Arici et~al.(2021)Arici, Seyfioglu, Neiman, Xu, Train, Chilimbi, Zeng,
  and Tutar]{arici2021mlim}
Tarik Arici, Mehmet~Saygin Seyfioglu, Tal Neiman, Yi~Xu, Son Train, Trishul
  Chilimbi, Belinda Zeng, and Ismail Tutar.
\newblock Mlim: Vision-and-language model pre-training with masked language and
  image modeling.
\newblock \emph{arXiv preprint arXiv:2109.12178}, 2021.

\bibitem[Ba et~al.(2016)Ba, Kiros, and Hinton]{ba2016layer}
Jimmy~Lei Ba, Jamie~Ryan Kiros, and Geoffrey~E Hinton.
\newblock Layer normalization.
\newblock \emph{arXiv preprint arXiv:1607.06450}, 2016.

\bibitem[Bachmann et~al.(2022)Bachmann, Mizrahi, Atanov, and
  Zamir]{bachmann2022multimae}
Roman Bachmann, David Mizrahi, Andrei Atanov, and Amir Zamir.
\newblock Multimae: Multi-modal multi-task masked autoencoders.
\newblock \emph{arXiv preprint arXiv:2204.01678}, 2022.

\bibitem[Bao et~al.(2021)Bao, Dong, and Wei]{bao2021beit}
Hangbo Bao, Li~Dong, and Furu Wei.
\newblock Beit: Bert pre-training of image transformers.
\newblock \emph{arXiv preprint arXiv:2106.08254}, 2021.

\bibitem[Bengio et~al.(2013)Bengio, Yao, Alain, and
  Vincent]{bengio2013generalized}
Yoshua Bengio, Li~Yao, Guillaume Alain, and Pascal Vincent.
\newblock Generalized denoising auto-encoders as generative models.
\newblock \emph{Advances in neural information processing systems}, 26, 2013.

\bibitem[Cai et~al.(2022)Cai, Kwon, Ravichandran, Bas, Tu, Bhotika, and
  Soatto]{cai2022x}
Zhaowei Cai, Gukyeong Kwon, Avinash Ravichandran, Erhan Bas, Zhuowen Tu, Rahul
  Bhotika, and Stefano Soatto.
\newblock X-detr: A versatile architecture for instance-wise vision-language
  tasks.
\newblock \emph{arXiv preprint arXiv:2204.05626}, 2022.

\bibitem[Chen et~al.(2020{\natexlab{a}})Chen, Radford, Child, Wu, Jun, Luan,
  and Sutskever]{chen2020generative}
Mark Chen, Alec Radford, Rewon Child, Jeffrey Wu, Heewoo Jun, David Luan, and
  Ilya Sutskever.
\newblock Generative pretraining from pixels.
\newblock In \emph{International Conference on Machine Learning}, pp.\
  1691--1703. PMLR, 2020{\natexlab{a}}.

\bibitem[Chen et~al.(2020{\natexlab{b}})Chen, Li, Yu, El~Kholy, Ahmed, Gan,
  Cheng, and Liu]{chen2020uniter}
Yen-Chun Chen, Linjie Li, Licheng Yu, Ahmed El~Kholy, Faisal Ahmed, Zhe Gan,
  Yu~Cheng, and Jingjing Liu.
\newblock Uniter: Universal image-text representation learning.
\newblock In \emph{European conference on computer vision}, pp.\  104--120.
  Springer, 2020{\natexlab{b}}.

\bibitem[Cho et~al.(2021)Cho, Lei, Tan, and Bansal]{cho2021unifying}
Jaemin Cho, Jie Lei, Hao Tan, and Mohit Bansal.
\newblock Unifying vision-and-language tasks via text generation.
\newblock In \emph{International Conference on Machine Learning}, pp.\
  1931--1942. PMLR, 2021.

\bibitem[Cubuk et~al.(2020)Cubuk, Zoph, Shlens, and Le]{cubuk2020randaugment}
Ekin~D Cubuk, Barret Zoph, Jonathon Shlens, and Quoc~V Le.
\newblock Randaugment: Practical automated data augmentation with a reduced
  search space.
\newblock In \emph{Proceedings of the IEEE/CVF Conference on Computer Vision
  and Pattern Recognition Workshops}, pp.\  702--703, 2020.

\bibitem[Deng et~al.(2009)Deng, Dong, Socher, Li, Li, and
  Fei-Fei]{deng2009imagenet}
Jia Deng, Wei Dong, Richard Socher, Li-Jia Li, Kai Li, and Li~Fei-Fei.
\newblock Imagenet: A large-scale hierarchical image database.
\newblock In \emph{2009 IEEE conference on computer vision and pattern
  recognition}, pp.\  248--255. Ieee, 2009.

\bibitem[Devlin et~al.(2018)Devlin, Chang, Lee, and Toutanova]{devlin2018bert}
Jacob Devlin, Ming-Wei Chang, Kenton Lee, and Kristina Toutanova.
\newblock Bert: Pre-training of deep bidirectional transformers for language
  understanding.
\newblock \emph{arXiv preprint arXiv:1810.04805}, 2018.

\bibitem[Dosovitskiy et~al.(2020)Dosovitskiy, Beyer, Kolesnikov, Weissenborn,
  Zhai, Unterthiner, Dehghani, Minderer, Heigold, Gelly,
  et~al.]{dosovitskiy2020image}
Alexey Dosovitskiy, Lucas Beyer, Alexander Kolesnikov, Dirk Weissenborn,
  Xiaohua Zhai, Thomas Unterthiner, Mostafa Dehghani, Matthias Minderer, Georg
  Heigold, Sylvain Gelly, et~al.
\newblock An image is worth 16x16 words: Transformers for image recognition at
  scale.
\newblock \emph{arXiv preprint arXiv:2010.11929}, 2020.

\bibitem[Dou et~al.(2022)Dou, Xu, Gan, Wang, Wang, Wang, Zhu, Zhang, Yuan,
  Peng, et~al.]{dou2022empirical}
Zi-Yi Dou, Yichong Xu, Zhe Gan, Jianfeng Wang, Shuohang Wang, Lijuan Wang,
  Chenguang Zhu, Pengchuan Zhang, Lu~Yuan, Nanyun Peng, et~al.
\newblock An empirical study of training end-to-end vision-and-language
  transformers.
\newblock In \emph{Proceedings of the IEEE/CVF Conference on Computer Vision
  and Pattern Recognition}, pp.\  18166--18176, 2022.

\bibitem[Duan et~al.(2022)Duan, Chen, Tran, Yang, Xu, Zeng, Tao, and
  Chilimbi]{duan2022multi}
Jiali Duan, Liqun Chen, Son Tran, Jinyu Yang, Yi~Xu, Belinda Zeng, Chenyang
  Tao, and Trishul Chilimbi.
\newblock Multi-modal alignment using representation codebook.
\newblock \emph{arXiv preprint arXiv:2203.00048}, 2022.

\bibitem[Fu et~al.(2021)Fu, Li, Gan, Lin, Wang, Wang, and Liu]{fu2021violet}
Tsu-Jui Fu, Linjie Li, Zhe Gan, Kevin Lin, William~Yang Wang, Lijuan Wang, and
  Zicheng Liu.
\newblock Violet: End-to-end video-language transformers with masked
  visual-token modeling.
\newblock \emph{arXiv preprint arXiv:2111.12681}, 2021.

\bibitem[Gan et~al.(2020)Gan, Chen, Li, Zhu, Cheng, and Liu]{gan2020large}
Zhe Gan, Yen-Chun Chen, Linjie Li, Chen Zhu, Yu~Cheng, and Jingjing Liu.
\newblock Large-scale adversarial training for vision-and-language
  representation learning.
\newblock \emph{Advances in Neural Information Processing Systems},
  33:\penalty0 6616--6628, 2020.

\bibitem[Geng et~al.(2022)Geng, Liu, Lee, Schuurams, Levine, and
  Abbeel]{geng2022multimodal}
Xinyang Geng, Hao Liu, Lisa Lee, Dale Schuurams, Sergey Levine, and Pieter
  Abbeel.
\newblock Multimodal masked autoencoders learn transferable representations.
\newblock \emph{arXiv preprint arXiv:2205.14204}, 2022.

\bibitem[Goyal et~al.(2017)Goyal, Khot, Summers-Stay, Batra, and
  Parikh]{goyal2017making}
Yash Goyal, Tejas Khot, Douglas Summers-Stay, Dhruv Batra, and Devi Parikh.
\newblock Making the v in vqa matter: Elevating the role of image understanding
  in visual question answering.
\newblock In \emph{Proceedings of the IEEE conference on computer vision and
  pattern recognition}, pp.\  6904--6913, 2017.

\bibitem[Griffin et~al.(2007)Griffin, Holub, and Perona]{griffin2007caltech}
Gregory Griffin, Alex Holub, and Pietro Perona.
\newblock Caltech-256 object category dataset.
\newblock 2007.

\bibitem[He et~al.(2020)He, Fan, Wu, Xie, and Girshick]{he2020momentum}
Kaiming He, Haoqi Fan, Yuxin Wu, Saining Xie, and Ross Girshick.
\newblock Momentum contrast for unsupervised visual representation learning.
\newblock In \emph{Proceedings of the IEEE/CVF conference on computer vision
  and pattern recognition}, pp.\  9729--9738, 2020.

\bibitem[He et~al.(2022)He, Chen, Xie, Li, Doll{\'a}r, and
  Girshick]{he2022masked}
Kaiming He, Xinlei Chen, Saining Xie, Yanghao Li, Piotr Doll{\'a}r, and Ross
  Girshick.
\newblock Masked autoencoders are scalable vision learners.
\newblock In \emph{Proceedings of the IEEE/CVF Conference on Computer Vision
  and Pattern Recognition}, pp.\  16000--16009, 2022.

\bibitem[Jia et~al.(2021)Jia, Yang, Xia, Chen, Parekh, Pham, Le, Sung, Li, and
  Duerig]{jia2021scaling}
Chao Jia, Yinfei Yang, Ye~Xia, Yi-Ting Chen, Zarana Parekh, Hieu Pham, Quoc Le,
  Yun-Hsuan Sung, Zhen Li, and Tom Duerig.
\newblock Scaling up visual and vision-language representation learning with
  noisy text supervision.
\newblock In \emph{International Conference on Machine Learning}, pp.\
  4904--4916. PMLR, 2021.

\bibitem[Kamath et~al.(2021)Kamath, Singh, LeCun, Synnaeve, Misra, and
  Carion]{kamath2021mdetr}
Aishwarya Kamath, Mannat Singh, Yann LeCun, Gabriel Synnaeve, Ishan Misra, and
  Nicolas Carion.
\newblock Mdetr-modulated detection for end-to-end multi-modal understanding.
\newblock In \emph{Proceedings of the IEEE/CVF International Conference on
  Computer Vision}, pp.\  1780--1790, 2021.

\bibitem[Karpathy \& Fei-Fei(2015)Karpathy and Fei-Fei]{karpathy2015deep}
Andrej Karpathy and Li~Fei-Fei.
\newblock Deep visual-semantic alignments for generating image descriptions.
\newblock In \emph{Proceedings of the IEEE conference on computer vision and
  pattern recognition}, pp.\  3128--3137, 2015.

\bibitem[Kim et~al.(2021)Kim, Son, and Kim]{kim2021vilt}
Wonjae Kim, Bokyung Son, and Ildoo Kim.
\newblock Vilt: Vision-and-language transformer without convolution or region
  supervision.
\newblock In \emph{International Conference on Machine Learning}, pp.\
  5583--5594. PMLR, 2021.

\bibitem[Krishna et~al.(2017)Krishna, Zhu, Groth, Johnson, Hata, Kravitz, Chen,
  Kalantidis, Li, Shamma, et~al.]{krishna2017visual}
Ranjay Krishna, Yuke Zhu, Oliver Groth, Justin Johnson, Kenji Hata, Joshua
  Kravitz, Stephanie Chen, Yannis Kalantidis, Li-Jia Li, David~A Shamma, et~al.
\newblock Visual genome: Connecting language and vision using crowdsourced
  dense image annotations.
\newblock \emph{International journal of computer vision}, 123\penalty0
  (1):\penalty0 32--73, 2017.

\bibitem[Li et~al.(2020{\natexlab{a}})Li, Duan, Fang, Gong, and
  Jiang]{li2020unicoder}
Gen Li, Nan Duan, Yuejian Fang, Ming Gong, and Daxin Jiang.
\newblock Unicoder-vl: A universal encoder for vision and language by
  cross-modal pre-training.
\newblock In \emph{Proceedings of the AAAI Conference on Artificial
  Intelligence}, pp.\  11336--11344, 2020{\natexlab{a}}.

\bibitem[Li et~al.(2021)Li, Selvaraju, Gotmare, Joty, Xiong, and
  Hoi]{li2021align}
Junnan Li, Ramprasaath Selvaraju, Akhilesh Gotmare, Shafiq Joty, Caiming Xiong,
  and Steven Chu~Hong Hoi.
\newblock Align before fuse: Vision and language representation learning with
  momentum distillation.
\newblock \emph{Advances in Neural Information Processing Systems}, 34, 2021.

\bibitem[Li et~al.(2022)Li, Li, Xiong, and Hoi]{li2022blip}
Junnan Li, Dongxu Li, Caiming Xiong, and Steven Hoi.
\newblock Blip: Bootstrapping language-image pre-training for unified
  vision-language understanding and generation.
\newblock \emph{arXiv preprint arXiv:2201.12086}, 2022.

\bibitem[Li et~al.(2019)Li, Yatskar, Yin, Hsieh, and Chang]{li2019visualbert}
Liunian~Harold Li, Mark Yatskar, Da~Yin, Cho-Jui Hsieh, and Kai-Wei Chang.
\newblock Visualbert: A simple and performant baseline for vision and language.
\newblock \emph{arXiv preprint arXiv:1908.03557}, 2019.

\bibitem[Li et~al.(2020{\natexlab{b}})Li, Yin, Li, Zhang, Hu, Zhang, Wang, Hu,
  Dong, Wei, et~al.]{li2020oscar}
Xiujun Li, Xi~Yin, Chunyuan Li, Pengchuan Zhang, Xiaowei Hu, Lei Zhang, Lijuan
  Wang, Houdong Hu, Li~Dong, Furu Wei, et~al.
\newblock Oscar: Object-semantics aligned pre-training for vision-language
  tasks.
\newblock In \emph{European Conference on Computer Vision}, pp.\  121--137.
  Springer, 2020{\natexlab{b}}.

\bibitem[Lin et~al.(2014)Lin, Maire, Belongie, Hays, Perona, Ramanan,
  Doll{\'a}r, and Zitnick]{lin2014microsoft}
Tsung-Yi Lin, Michael Maire, Serge Belongie, James Hays, Pietro Perona, Deva
  Ramanan, Piotr Doll{\'a}r, and C~Lawrence Zitnick.
\newblock Microsoft coco: Common objects in context.
\newblock In \emph{European conference on computer vision}, pp.\  740--755.
  Springer, 2014.

\bibitem[Liu et~al.(2019)Liu, Ott, Goyal, Du, Joshi, Chen, Levy, Lewis,
  Zettlemoyer, and Stoyanov]{liu2019roberta}
Yinhan Liu, Myle Ott, Naman Goyal, Jingfei Du, Mandar Joshi, Danqi Chen, Omer
  Levy, Mike Lewis, Luke Zettlemoyer, and Veselin Stoyanov.
\newblock Roberta: A robustly optimized bert pretraining approach.
\newblock \emph{arXiv preprint arXiv:1907.11692}, 2019.

\bibitem[Loshchilov \& Hutter(2017)Loshchilov and
  Hutter]{loshchilov2017decoupled}
Ilya Loshchilov and Frank Hutter.
\newblock Decoupled weight decay regularization.
\newblock \emph{arXiv preprint arXiv:1711.05101}, 2017.

\bibitem[Lu et~al.(2019)Lu, Batra, Parikh, and Lee]{lu2019vilbert}
Jiasen Lu, Dhruv Batra, Devi Parikh, and Stefan Lee.
\newblock Vilbert: Pretraining task-agnostic visiolinguistic representations
  for vision-and-language tasks.
\newblock \emph{Advances in neural information processing systems}, 32, 2019.

\bibitem[Lu et~al.(2020)Lu, Goswami, Rohrbach, Parikh, and Lee]{lu202012}
Jiasen Lu, Vedanuj Goswami, Marcus Rohrbach, Devi Parikh, and Stefan Lee.
\newblock 12-in-1: Multi-task vision and language representation learning.
\newblock In \emph{Proceedings of the IEEE/CVF Conference on Computer Vision
  and Pattern Recognition}, pp.\  10437--10446, 2020.

\bibitem[Mokady et~al.(2021)Mokady, Hertz, and Bermano]{mokady2021clipcap}
Ron Mokady, Amir Hertz, and Amit~H Bermano.
\newblock Clipcap: Clip prefix for image captioning.
\newblock \emph{arXiv preprint arXiv:2111.09734}, 2021.

\bibitem[Nilsback \& Zisserman(2008)Nilsback and
  Zisserman]{nilsback2008automated}
Maria-Elena Nilsback and Andrew Zisserman.
\newblock Automated flower classification over a large number of classes.
\newblock In \emph{2008 Sixth Indian Conference on Computer Vision, Graphics \&
  Image Processing}, pp.\  722--729. IEEE, 2008.

\bibitem[Ordonez et~al.(2011)Ordonez, Kulkarni, and Berg]{ordonez2011im2text}
Vicente Ordonez, Girish Kulkarni, and Tamara Berg.
\newblock Im2text: Describing images using 1 million captioned photographs.
\newblock \emph{Advances in neural information processing systems}, 24, 2011.

\bibitem[Peng et~al.(2022)Peng, Dong, Bao, Ye, and Wei]{peng2022beit}
Zhiliang Peng, Li~Dong, Hangbo Bao, Qixiang Ye, and Furu Wei.
\newblock Beit v2: Masked image modeling with vector-quantized visual
  tokenizers.
\newblock \emph{arXiv preprint arXiv:2208.06366}, 2022.

\bibitem[Plummer et~al.(2015)Plummer, Wang, Cervantes, Caicedo, Hockenmaier,
  and Lazebnik]{plummer2015flickr30k}
Bryan~A Plummer, Liwei Wang, Chris~M Cervantes, Juan~C Caicedo, Julia
  Hockenmaier, and Svetlana Lazebnik.
\newblock Flickr30k entities: Collecting region-to-phrase correspondences for
  richer image-to-sentence models.
\newblock In \emph{Proceedings of the IEEE international conference on computer
  vision}, pp.\  2641--2649, 2015.

\bibitem[Qi et~al.(2020)Qi, Su, Song, Cui, Bharti, and
  Sacheti]{qi2020imagebert}
Di~Qi, Lin Su, Jia Song, Edward Cui, Taroon Bharti, and Arun Sacheti.
\newblock Imagebert: Cross-modal pre-training with large-scale weak-supervised
  image-text data.
\newblock \emph{arXiv preprint arXiv:2001.07966}, 2020.

\bibitem[Quattoni \& Torralba(2009)Quattoni and
  Torralba]{quattoni2009recognizing}
Ariadna Quattoni and Antonio Torralba.
\newblock Recognizing indoor scenes.
\newblock In \emph{2009 IEEE conference on computer vision and pattern
  recognition}, pp.\  413--420. IEEE, 2009.

\bibitem[Radford et~al.(2018)Radford, Narasimhan, Salimans, and
  Sutskever]{radford2018improving}
Alec Radford, Karthik Narasimhan, Tim Salimans, and Ilya Sutskever.
\newblock Improving language understanding by generative pre-training.
\newblock 2018.

\bibitem[Radford et~al.(2019)Radford, Wu, Child, Luan, Amodei, Sutskever,
  et~al.]{radford2019language}
Alec Radford, Jeffrey Wu, Rewon Child, David Luan, Dario Amodei, Ilya
  Sutskever, et~al.
\newblock Language models are unsupervised multitask learners.
\newblock \emph{OpenAI blog}, 1\penalty0 (8):\penalty0 9, 2019.

\bibitem[Radford et~al.(2021)Radford, Kim, Hallacy, Ramesh, Goh, Agarwal,
  Sastry, Askell, Mishkin, Clark, et~al.]{radford2021learning}
Alec Radford, Jong~Wook Kim, Chris Hallacy, Aditya Ramesh, Gabriel Goh,
  Sandhini Agarwal, Girish Sastry, Amanda Askell, Pamela Mishkin, Jack Clark,
  et~al.
\newblock Learning transferable visual models from natural language
  supervision.
\newblock In \emph{International Conference on Machine Learning}, pp.\
  8748--8763. PMLR, 2021.

\bibitem[Ramesh et~al.(2021)Ramesh, Pavlov, Goh, Gray, Voss, Radford, Chen, and
  Sutskever]{ramesh2021zero}
Aditya Ramesh, Mikhail Pavlov, Gabriel Goh, Scott Gray, Chelsea Voss, Alec
  Radford, Mark Chen, and Ilya Sutskever.
\newblock Zero-shot text-to-image generation.
\newblock In \emph{International Conference on Machine Learning}, pp.\
  8821--8831. PMLR, 2021.

\bibitem[Ramesh et~al.(2022)Ramesh, Dhariwal, Nichol, Chu, and
  Chen]{ramesh2022hierarchical}
Aditya Ramesh, Prafulla Dhariwal, Alex Nichol, Casey Chu, and Mark Chen.
\newblock Hierarchical text-conditional image generation with clip latents.
\newblock \emph{arXiv preprint arXiv:2204.06125}, 2022.

\bibitem[Sharma et~al.(2018)Sharma, Ding, Goodman, and
  Soricut]{sharma2018conceptual}
Piyush Sharma, Nan Ding, Sebastian Goodman, and Radu Soricut.
\newblock Conceptual captions: A cleaned, hypernymed, image alt-text dataset
  for automatic image captioning.
\newblock In \emph{Proceedings of the 56th Annual Meeting of the Association
  for Computational Linguistics (Volume 1: Long Papers)}, pp.\  2556--2565,
  2018.

\bibitem[Shen et~al.(2021)Shen, Li, Tan, Bansal, Rohrbach, Chang, Yao, and
  Keutzer]{shen2021much}
Sheng Shen, Liunian~Harold Li, Hao Tan, Mohit Bansal, Anna Rohrbach, Kai-Wei
  Chang, Zhewei Yao, and Kurt Keutzer.
\newblock How much can clip benefit vision-and-language tasks?
\newblock \emph{arXiv preprint arXiv:2107.06383}, 2021.

\bibitem[Singh et~al.(2022)Singh, Hu, Goswami, Couairon, Galuba, Rohrbach, and
  Kiela]{singh2022flava}
Amanpreet Singh, Ronghang Hu, Vedanuj Goswami, Guillaume Couairon, Wojciech
  Galuba, Marcus Rohrbach, and Douwe Kiela.
\newblock Flava: A foundational language and vision alignment model.
\newblock In \emph{Proceedings of the IEEE/CVF Conference on Computer Vision
  and Pattern Recognition}, pp.\  15638--15650, 2022.

\bibitem[Sohn et~al.(2014)Sohn, Shang, and Lee]{sohn2014improved}
Kihyuk Sohn, Wenling Shang, and Honglak Lee.
\newblock Improved multimodal deep learning with variation of information.
\newblock \emph{Advances in neural information processing systems}, 27, 2014.

\bibitem[Su et~al.(2019)Su, Zhu, Cao, Li, Lu, Wei, and Dai]{su2019vl}
Weijie Su, Xizhou Zhu, Yue Cao, Bin Li, Lewei Lu, Furu Wei, and Jifeng Dai.
\newblock Vl-bert: Pre-training of generic visual-linguistic representations.
\newblock \emph{arXiv preprint arXiv:1908.08530}, 2019.

\bibitem[Suhr et~al.(2018)Suhr, Zhou, Zhang, Zhang, Bai, and
  Artzi]{suhr2018corpus}
Alane Suhr, Stephanie Zhou, Ally Zhang, Iris Zhang, Huajun Bai, and Yoav Artzi.
\newblock A corpus for reasoning about natural language grounded in
  photographs.
\newblock \emph{arXiv preprint arXiv:1811.00491}, 2018.

\bibitem[Tan \& Bansal(2019)Tan and Bansal]{tan2019lxmert}
Hao Tan and Mohit Bansal.
\newblock Lxmert: Learning cross-modality encoder representations from
  transformers.
\newblock \emph{arXiv preprint arXiv:1908.07490}, 2019.

\bibitem[Tsimpoukelli et~al.(2021)Tsimpoukelli, Menick, Cabi, Eslami, Vinyals,
  and Hill]{tsimpoukelli2021multimodal}
Maria Tsimpoukelli, Jacob~L Menick, Serkan Cabi, SM~Eslami, Oriol Vinyals, and
  Felix Hill.
\newblock Multimodal few-shot learning with frozen language models.
\newblock \emph{Advances in Neural Information Processing Systems},
  34:\penalty0 200--212, 2021.

\bibitem[Vaswani et~al.(2017)Vaswani, Shazeer, Parmar, Uszkoreit, Jones, Gomez,
  Kaiser, and Polosukhin]{vaswani2017attention}
Ashish Vaswani, Noam Shazeer, Niki Parmar, Jakob Uszkoreit, Llion Jones,
  Aidan~N Gomez, {\L}ukasz Kaiser, and Illia Polosukhin.
\newblock Attention is all you need.
\newblock \emph{Advances in neural information processing systems}, 30, 2017.

\bibitem[Wah et~al.(2011)Wah, Branson, Welinder, Perona, and
  Belongie]{wah2011caltech}
Catherine Wah, Steve Branson, Peter Welinder, Pietro Perona, and Serge
  Belongie.
\newblock The caltech-ucsd birds-200-2011 dataset.
\newblock 2011.

\bibitem[Wang et~al.(2022)Wang, Bao, Dong, Bjorck, Peng, Liu, Aggarwal,
  Mohammed, Singhal, Som, et~al.]{wang2022image}
Wenhui Wang, Hangbo Bao, Li~Dong, Johan Bjorck, Zhiliang Peng, Qiang Liu, Kriti
  Aggarwal, Owais~Khan Mohammed, Saksham Singhal, Subhojit Som, et~al.
\newblock Image as a foreign language: Beit pretraining for all vision and
  vision-language tasks.
\newblock \emph{arXiv preprint arXiv:2208.10442}, 2022.

\bibitem[Wang et~al.(2021)Wang, Yu, Yu, Dai, Tsvetkov, and Cao]{wang2021simvlm}
Zirui Wang, Jiahui Yu, Adams~Wei Yu, Zihang Dai, Yulia Tsvetkov, and Yuan Cao.
\newblock Simvlm: Simple visual language model pretraining with weak
  supervision.
\newblock \emph{arXiv preprint arXiv:2108.10904}, 2021.

\bibitem[Wettig et~al.(2022)Wettig, Gao, Zhong, and Chen]{wettig2022should}
Alexander Wettig, Tianyu Gao, Zexuan Zhong, and Danqi Chen.
\newblock Should you mask 15\% in masked language modeling?
\newblock \emph{arXiv preprint arXiv:2202.08005}, 2022.

\bibitem[Wightman(2019)]{rw2019timm}
Ross Wightman.
\newblock Pytorch image models.
\newblock \url{https://github.com/rwightman/pytorch-image-models}, 2019.

\bibitem[Wolf et~al.(2020)Wolf, Debut, Sanh, Chaumond, Delangue, Moi, Cistac,
  Rault, Louf, Funtowicz, Davison, Shleifer, von Platen, Ma, Jernite, Plu, Xu,
  Scao, Gugger, Drame, Lhoest, and Rush]{wolf-etal-2020-transformers}
Thomas Wolf, Lysandre Debut, Victor Sanh, Julien Chaumond, Clement Delangue,
  Anthony Moi, Pierric Cistac, Tim Rault, Rémi Louf, Morgan Funtowicz, Joe
  Davison, Sam Shleifer, Patrick von Platen, Clara Ma, Yacine Jernite, Julien
  Plu, Canwen Xu, Teven~Le Scao, Sylvain Gugger, Mariama Drame, Quentin Lhoest,
  and Alexander~M. Rush.
\newblock Transformers: State-of-the-art natural language processing.
\newblock In \emph{Proceedings of the 2020 Conference on Empirical Methods in
  Natural Language Processing: System Demonstrations}, pp.\  38--45, Online,
  October 2020. Association for Computational Linguistics.
\newblock URL \url{https://www.aclweb.org/anthology/2020.emnlp-demos.6}.

\bibitem[Xie et~al.(2019)Xie, Lai, Doran, and Kadav]{xie2019visual}
Ning Xie, Farley Lai, Derek Doran, and Asim Kadav.
\newblock Visual entailment: A novel task for fine-grained image understanding.
\newblock \emph{arXiv preprint arXiv:1901.06706}, 2019.

\bibitem[Xie et~al.(2021)Xie, Zhang, Cao, Lin, Bao, Yao, Dai, and
  Hu]{xie2021simmim}
Zhenda Xie, Zheng Zhang, Yue Cao, Yutong Lin, Jianmin Bao, Zhuliang Yao,
  Qi~Dai, and Han Hu.
\newblock Simmim: A simple framework for masked image modeling.
\newblock \emph{arXiv preprint arXiv:2111.09886}, 2021.

\bibitem[Yang et~al.(2022)Yang, Duan, Tran, Xu, Chanda, Chen, Zeng, Chilimbi,
  and Huang]{yang2022vision}
Jinyu Yang, Jiali Duan, Son Tran, Yi~Xu, Sampath Chanda, Liqun Chen, Belinda
  Zeng, Trishul Chilimbi, and Junzhou Huang.
\newblock Vision-language pre-training with triple contrastive learning.
\newblock \emph{arXiv preprint arXiv:2202.10401}, 2022.

\bibitem[Yang \& Newsam(2010)Yang and Newsam]{yang2010bag}
Yi~Yang and Shawn Newsam.
\newblock Bag-of-visual-words and spatial extensions for land-use
  classification.
\newblock In \emph{Proceedings of the 18th SIGSPATIAL international conference
  on advances in geographic information systems}, pp.\  270--279, 2010.

\bibitem[Yang et~al.(2019)Yang, Dai, Yang, Carbonell, Salakhutdinov, and
  Le]{yang2019xlnet}
Zhilin Yang, Zihang Dai, Yiming Yang, Jaime Carbonell, Russ~R Salakhutdinov,
  and Quoc~V Le.
\newblock Xlnet: Generalized autoregressive pretraining for language
  understanding.
\newblock \emph{Advances in neural information processing systems}, 32, 2019.

\bibitem[Yu et~al.(2022)Yu, Wang, Vasudevan, Yeung, Seyedhosseini, and
  Wu]{yu2022coca}
Jiahui Yu, Zirui Wang, Vijay Vasudevan, Legg Yeung, Mojtaba Seyedhosseini, and
  Yonghui Wu.
\newblock Coca: Contrastive captioners are image-text foundation models.
\newblock \emph{arXiv preprint arXiv:2205.01917}, 2022.

\bibitem[Yuan et~al.(2021)Yuan, Chen, Chen, Codella, Dai, Gao, Hu, Huang, Li,
  Li, et~al.]{yuan2021florence}
Lu~Yuan, Dongdong Chen, Yi-Ling Chen, Noel Codella, Xiyang Dai, Jianfeng Gao,
  Houdong Hu, Xuedong Huang, Boxin Li, Chunyuan Li, et~al.
\newblock Florence: A new foundation model for computer vision.
\newblock \emph{arXiv preprint arXiv:2111.11432}, 2021.

\bibitem[Zhang et~al.(2021)Zhang, Li, Hu, Yang, Zhang, Wang, Choi, and
  Gao]{zhang2021vinvl}
Pengchuan Zhang, Xiujun Li, Xiaowei Hu, Jianwei Yang, Lei Zhang, Lijuan Wang,
  Yejin Choi, and Jianfeng Gao.
\newblock Vinvl: Revisiting visual representations in vision-language models.
\newblock In \emph{Proceedings of the IEEE/CVF Conference on Computer Vision
  and Pattern Recognition}, pp.\  5579--5588, 2021.

\end{thebibliography}
\bibliographystyle{iclr2023_conference}

\newpage
\appendix
\section{Appendix}\label{sec:appendix}
\subsection{Details on finetuning for downstream tasks}

We explain implementation details for each of the downstream tasks. For all the downstream tasks, we use AdamW~\citep{loshchilov2017decoupled} with a weight decay of 0.05 and the cosine scheduler. An image size of $384 \times 384$ with RandAugment~\citep{cubuk2020randaugment} is utilized and the positional encoding is interpolated following~\citep{dosovitskiy2020image}. Except for the VQA task, we use the model achieves the best performance in the validation set to report the performance on the test set. We use the last epoch model for the VQA evaluation.

\textbf{Image-Text Retrieval: }COCO~\citep{lin2014microsoft} and Flickr30k~\citep{plummer2015flickr30k} are used to report the performance. To be specific, we follow data splits proposed in~\citep{karpathy2015deep} and an average recall over image and text retrieval is used to find the best model in the validation set. The pre-trained model is finetuned for 15 epochs with a batch size of 256 and a learning rate of $1\times 10^{-5}$. 

\textbf{Visual Question Answering (VQA): }For a fair comparison with existing methods~\citep{chen2020uniter, li2021align}, we use training and validation sets from VQA v2.0~\citep{goyal2017making} with a subset of VQA samples from Visual Genome~\citep{krishna2017visual} for training. Also, we report performance on both test-dev and test-std splits of VQA v2.0. Following~\citep{li2021align}, we weight the loss for each answer based on its occurrence among all the answers. The model is finetuned for 15 epochs with a batch size of 256 . We use a learning rate of $2 \times 10^{-5}$ for the image and the text cross-modality encoders, the fusion encoder, and the answer decoder. For the image and the text encoders, a learning rate of $1 \times 10^{-5}$ is used. The fusion encoder and the answer decoder are initialized by the last and all three blocks of the pre-trained text cross-modality encoder, respectively. 

\textbf{Natural Language for Visual Reasoning (NLVR): }Data splits proposed in~\citep{suhr2018corpus} are used for finetuning and evaluation. The model is finetuned for 5 epochs with a batch size of 128. Since the classifier is newly added after finetuning, we use a learning rate of $1 \times 10^{-4}$ for the classifier and $1 \times 10^{-5}$ for the remaining parts of the model. Different from~\citep{duan2022multi, li2021align, yang2022vision}, where the models require additional text-assignment pre-training step before finetuning, we directly finetune for simplicity.

\textbf{Visual Entailment (VE): } We follow data splits proposed in SNLI-VE~\citep{xie2019visual}. We finetune the model with a batch size of 256 for 5 epochs. Similar to the NLVR task, a learning rate of $1 \times 10^{-4}$ is used for the classifier and $1 \times 10^{-5}$ is used for the remaining parts of the model.

\begin{figure}[h]
\centering
\begin{minipage}[t]{0.4\linewidth}
  \centering
  \subcaptionbox{VQA}{\includegraphics[width=\linewidth]{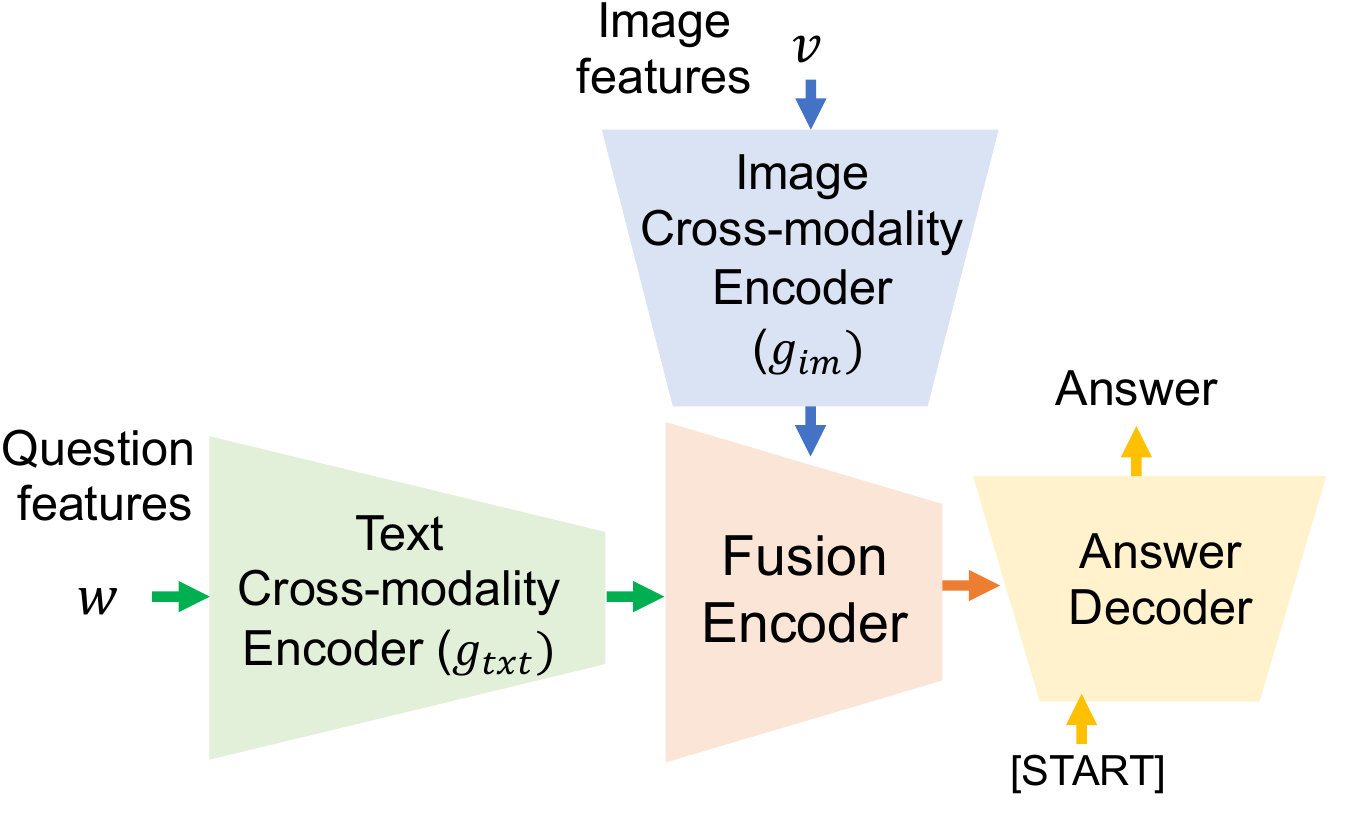}}
\end{minipage}
\hspace{0.5cm}
\begin{minipage}[t]{0.4\linewidth}
  \centering
  \subcaptionbox{NLVR}{\includegraphics[width=\linewidth]{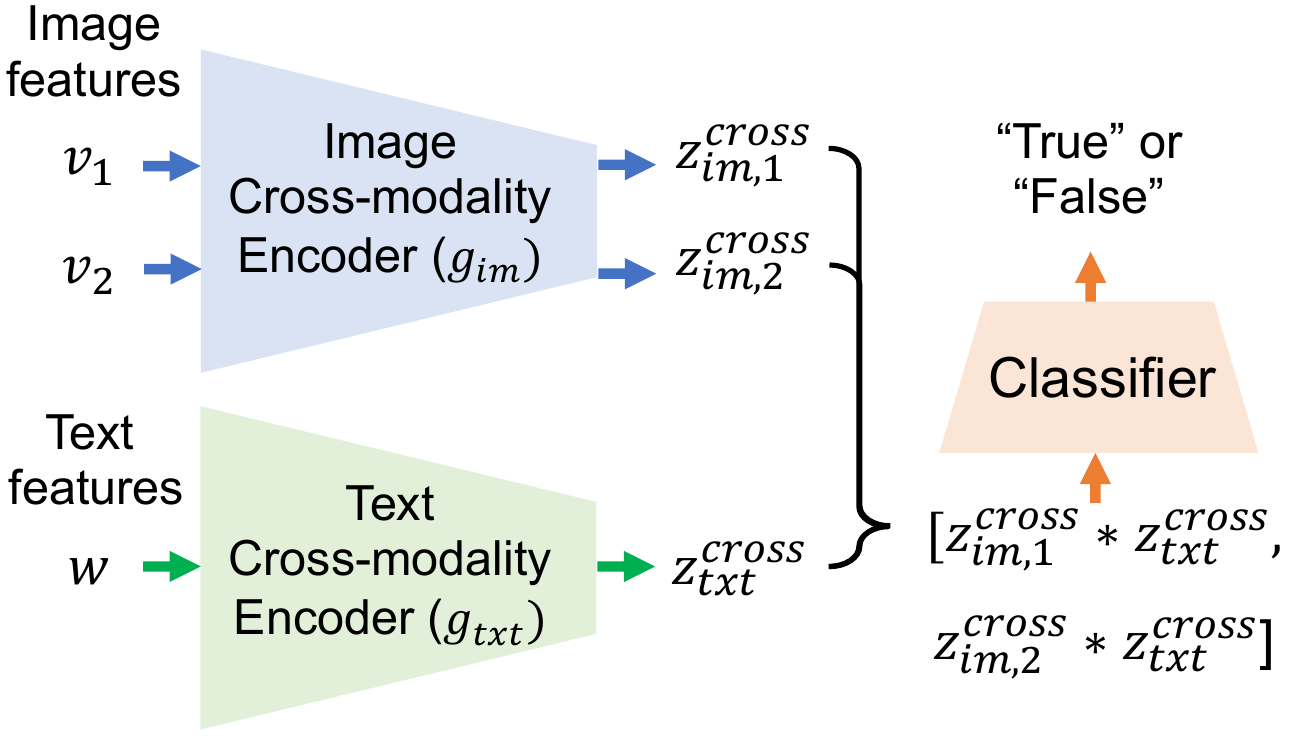}}
\end{minipage}\caption{An illustration of model architectures for VQA and NLVR.}\label{fig:vqa_nlvr}
\end{figure}

\subsection{Reproducibility}
\begin{figure}[t]
    \centering
    \includegraphics[width=.75\textwidth]{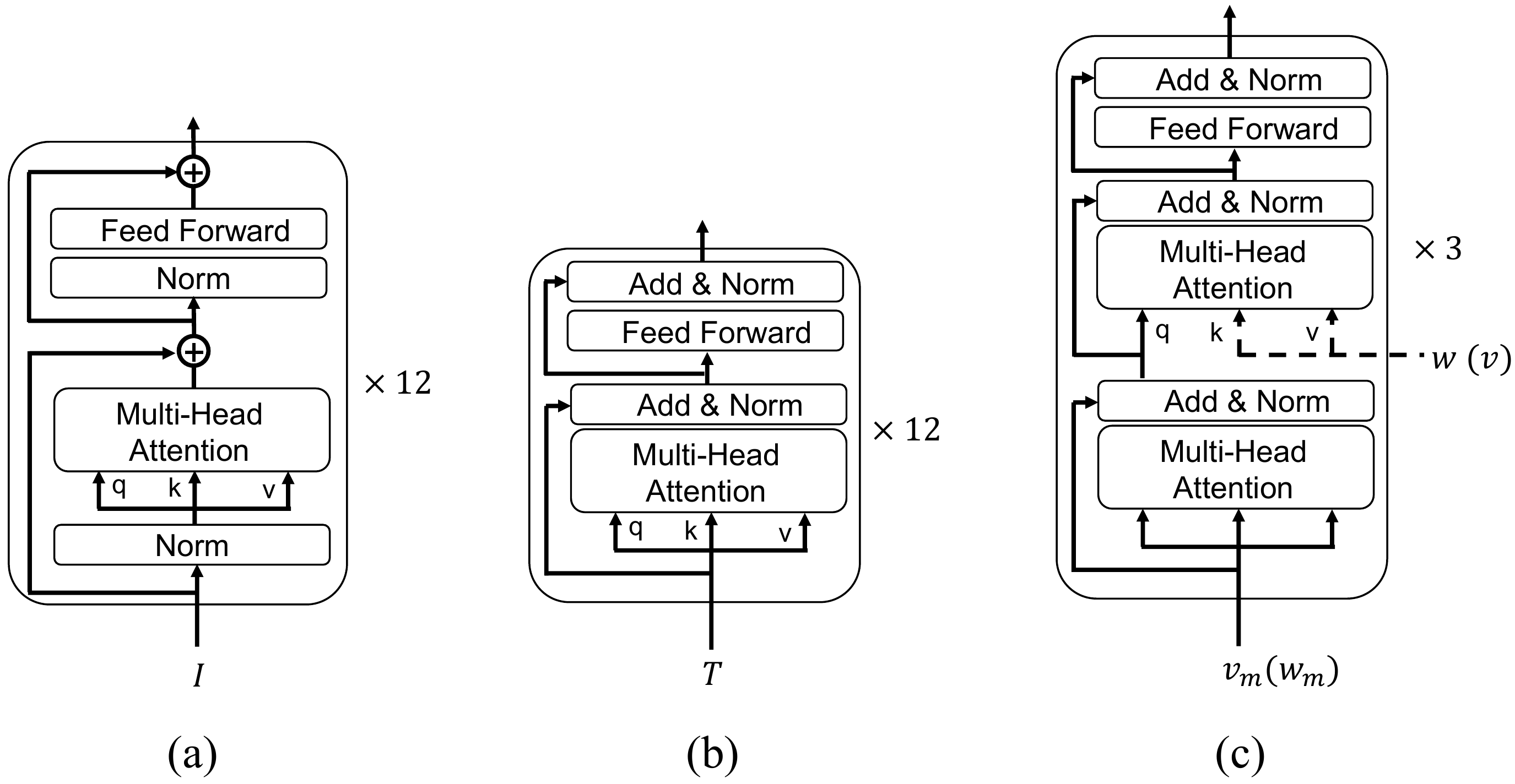}
    \caption{Model architectures of (a) Image encoder, (b) Text encoder and (c) Image (text) cross-modality encoder. The dotted lines in (c) denote key and value from the other modality for cross-attention.}\label{fig:appendix_blocks}
\end{figure}

We add more details of MaskVLM for reproducibility. We used the ImageNet pre-trained ViT (\texttt{vit\_base\_patch16\_224}) from~\citep{rw2019timm} and the pre-trained RoBERTa (\texttt{roberta-base}) from Hugging Face~\citep{wolf-etal-2020-transformers}. The detailed model architectures are visualized in Figure~\ref{fig:appendix_blocks}. Following~\citep{dosovitskiy2020image}, the image encoder uses layer normalization~\citep{ba2016layer} before each multi-head attention block while the text encoder applies layer normalization after each multi-head attention block (post norm). For the image (text) cross-modality encoder, we adopt the post norm and use the outputs of the text (image) encoder as keys and values to compute cross-attention. To compute MIM and MLM, the self-attention outputs of the masked image features, $v_m$, is used as queries and the unmasked text features, $w$, are used as keys and values in the image cross-modality encoder. For the text cross-modality encoder, the masked text features, $w_m$, are used as queries and the unmasked image features, $v$, are used as keys and values. To keep the framework simple, we do not use any loss weighting for each loss term and layer decay during finetuning.

\subsection{Ablation study on masking strategies}
\begin{table}[t]
\aboverulesep = 0.48mm
\belowrulesep = 0.48mm
\small
\centering
\setlength\tabcolsep{4pt}
\resizebox{\textwidth}{!}{%
\begin{tabular}{ccccccc|cccccc}
\toprule
\multirow{3}{*}{Method} & \multicolumn{6}{c}{MSCOCO (5K)} & \multicolumn{6}{c}{Flickr30k (1K)} \\
 & \multicolumn{3}{c}{Image Retrieval} & \multicolumn{3}{c}{Text Retrieval} & \multicolumn{3}{c}{Image Retrieval} & \multicolumn{3}{c}{Text Retrieval} \\ 
 & R@1 & R@5 & R@10 & R@1 & R@5 & R@10 & R@1 & R@5 & R@10 & R@1 & R@5 & R@10 \\ \midrule
ALBEF & 56.8 & 81.5 & 89.2 & 73.1 & 91.4 & 96.0 & 82.8 & 96.7 & 98.4 & 94.3 & 99.4 & 99.8 \\
MaskVLM (both) & \textbf{59.5} & 83.4 & 90.2 & \textbf{76.0} & 93.4 & 96.9 & \textbf{83.9} & 96.6 & 98.3 & \textbf{95.1} & 99.9 & 100.0 \\
MaskVLM (one) & \textbf{60.1} & 83.6 & 90.4 & \textbf{76.3} & 93.8 & 96.8 & \textbf{84.5} & 96.7 & 98.2 & \textbf{95.6} & 99.4 & 99.9 \\
\bottomrule
\end{tabular}\vspace{0.3cm}\caption{Comparison of finetuned MaskVLMs with different masking strategies and ALBEF on image-text retrieval. (MaskVLM (One): masking one modality at a time for computing MLM and MIM losses. MaskVLM (both): masking both modalities at the same time for reconstruction)}\label{tab:masking_strategy}}
\end{table}
We study different masking strategies in computing MIM and MLM losses. In particular, we compare MaskVLM using one modality masked and the other modality unmasked for reconstruction (MaskVLM (one)) and MaskVLM using both modalities masked at the same time for reconstruction. We compare these two MaskVLM models with the state-of-the-art method, ALBEF in Table~\ref{tab:masking_strategy}. We follow the experimental setup described in Section~\ref{subsec:datasets} and report the finetuning performance on image-text retrieval. The performance of masking one modality at a time (MaskVLM (one)) was slightly better than masking both modalities at the same time (MaskVLM (both)). However, we observed that both reconstruction strategies are still effective as they achieve higher R@1 for image and text retrieval in both COCO and Flickr30k compared to ALBEF.  

\subsection{Ablation study on masking ratio}
We perform ablation study using different masking ratios for masked vision and language modeling. In particular, we pre-train MaskVLM with several combinations of image and text masking ratios on the CC 50\% + COCO dataset and report the finetuned image-text retrieval performance on Flickr30k in Table~\ref{tab:ablation_masking_ratio}. We also report an average of R@k for image and text retrieval. When only image masking ratio is changed in the first three rows of the table, the difference between the maximum and the minimum of the average recall is 0.26 for image retrieval and 0.10 for text retrieval. This shows that MaskVLM achieves stable performance across the tested image masking ratios. From comparison between the second row and the last row, we observe that increasing the text masking ratio from 0.15 to 0.3 leads to higher recall performance for both image and text retrieval.

\begin{table}[h]
\begin{tabular}{ccccc|cccc}
\toprule
\multirow{2}{*}{\begin{tabular}[c]{@{}c@{}}Masking ratio\\ (image / text)\end{tabular}} & \multicolumn{4}{c}{Image Retrieval} & \multicolumn{4}{c}{Text Retrieval} \\ 
 & R@1 & R@5 & R@10 & Average & R@1 & R@5 & R@10 & Average \\ \midrule
0.5 / 0.3 & 81.32 & 96.04 & 97.92 & 91.76 & 93.30 & 99.70 & 100.00 & 97.67 \\
0.6 / 0.3 & 81.26 & 96.00 & 97.78 & 91.68 & 94.10 & 99.60 & 99.60 & 97.77 \\
0.7 / 0.3 & 81.82 & 96.00 & 98.00 & 91.94 & 93.60 & 99.50 & 99.90 & 97.67 \\
0.6 / 0.15 & 80.30 & 95.66 & 97.82 & 91.26 & 92.50 & 99.10 & 99.60 & 97.07 \\
\bottomrule
\end{tabular}\vspace{0.2cm}\caption{Finetuned image-text retrieval performance on Flickr30k with different masking ratios for masked vision and language modeling.}\label{tab:ablation_masking_ratio}
\end{table}

\subsection{Evaluation on image recognition}
We evaluate the image recognition performance of MaskVLM. Following CLIP~\citep{radford2021learning}, we perform image classification directly using the pre-trained MaskVLM on various image recognition datasets including UC Merced Land Use~\citep{yang2010bag}, MIT-67~\citep{quattoni2009recognizing}, CUB-200~\citep{wah2011caltech}, Oxford Flowers~\citep{nilsback2008automated}, Caltech-256~\citep{griffin2007caltech}, and ImageNet-1K~\citep{deng2009imagenet}. We compare the Top-1 accuracy of MaskVLM with ALBEF~\citep{li2021align} in Table~\ref{tab:image_recognition}. Both models are pre-trained with the 4M dataset. Since during pre-training stage, both MaskVLM and ALBEF were initialized with the ImageNet pre-trained weights, the evaluation on ImageNet is not strictly zero-shot but the evaluation on other datasets is. We formulate image classification as image-to-text retrieval where the similarity scores between a query image and all the class names are computed to retrieve top-1 class name. The similarity scores can be obtained using either ITC and ITM scores, and we report them separately. Also, the results of using prompt engineering as in CLIP are reported.

As shown in Table~\ref{tab:image_recognition}, MaskVLM consistently outperforms ALBEF across all the datasets. In particular, prompt engineering improves the average accuracy across all the datasets for MaskVLM but ALBEF achieves lower average accuracy with prompt engineering. This shows that MaskVLM can better align images with variants of text than ALBEF, which results in stronger V+L representations of MaskVLM for the image recognition task.

\begin{table}[h]
\aboverulesep = 0.48mm
\belowrulesep = 0.48mm
\small
\centering
\setlength\tabcolsep{4pt}
\resizebox{\textwidth}{!}{%
\begin{tabular}{ccccccccc}
\toprule
Method & \begin{tabular}[c]{@{}c@{}}Prompt\\ Engineering\end{tabular} & \begin{tabular}[c]{@{}c@{}}UC Merced\\ Land Use\end{tabular} & MIT-67 & CUB-200 & \begin{tabular}[c]{@{}c@{}}Oxford \\ Flowers\end{tabular} & Caltech256 & ImageNet-1K & Average \\ \bottomrule
ALBEF (ITC) & \xmark & 31.62 & 52.46 & 5.78 & 26.93 & 45.75 & 36.28 & 33.14\\
ALBEF (ITM) & \xmark & 26.29 & 56.19 & 5.68 & 21.68 & 41.79 & 35.41 & 31.17 \\
ALBEF (ITC) & \cmark & 28.38 & 51.04 & 5.33 & 25.44 & 48.71 & 30.26 & 31.53 \\
ALBEF (ITM) & \cmark & 15.81 & 29.70 & 3.90 & 18.86 & 26.57 & 12.53 & 17.90 \\ \midrule
MaskVLM (ITC) & \xmark & 41.52 & 55.60 & 4.97 & 26.61 & 60.68 & 33.16 & 37.09\\
MaskVLM (ITM) & \xmark & 29.52 & 58.21 & \textbf{10.36} & 36.14 & 58.95 & 34.59 & 37.96 \\
MaskVLM (ITC) & \cmark & \textbf{45.14} & \textbf{66.27} & 4.37 & 32.72 & 60.95 & 39.16 & 41.43 \\
MaskVLM (ITM) & \cmark & 40.76 & 63.36 & 8.53 & \textbf{44.53} & \textbf{61.77} & \textbf{39.71} & \textbf{43.11} \\ \bottomrule
\end{tabular}}\caption{Top-1 accuracy of pre-trained MaskVLM and ALBEF on image recognition. ITC and ITM denote the alignment scores utilized to perform image classification.}\label{tab:image_recognition}
\end{table}

\subsection{Additional examples for the qualitative analysis}

\begin{figure}[ht]
    \centering
    \includegraphics[width=\textwidth]{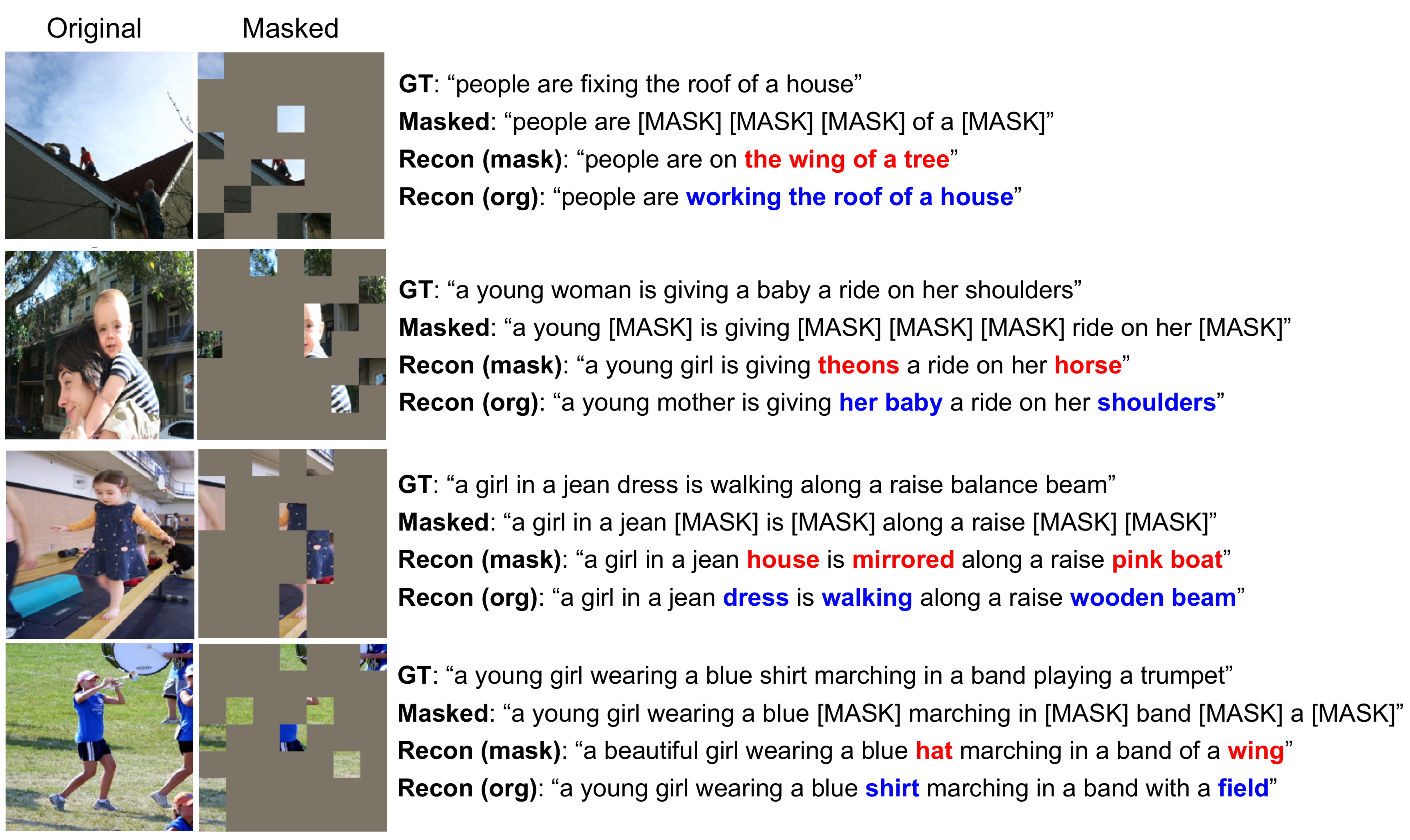}
    \vspace{0.1cm}\caption{Additional masked language modeling examples using masked and original images. ``Recon (mask)'' and ``Recon (org)'' denote reconstructed text using the masked image and the original image, respectively.}\label{fig:add_example}
\end{figure}

We present additional examples for the qualitative analysis of MaskVLM in Figure~\ref{fig:add_example}. Similar to Figure~\ref{fig:example}, masked text tokens are reconstructed with masked images (``Recon (mask)'') and original images (``Recon (org)''). We highlight that MaskVLM utilizes both V+L information to reconstruct the text which corresponds to the given image. 

\subsection{Statistics of the pre-training dataset}

In Table~\ref{tab:dataset_stat}, we report the statistics of the 4M pre-training dataset that MaskVLM is trained on. We note that some data urls provided in the web datasets can become invalid, which may lead to slightly different number of image-text pairs depending on when the datasets are downloaded. 

\begin{table}[ht]
\small
\centering
\begin{tabular}{ccc}
    \toprule
    Dataset & \# of image-text pairs & \# of images \\ \midrule
    COCO & 566,747 & 113,287 \\
    CC & 2,912,317 & 2,912,317 \\
    SBU & 1,000,000 & 1,000,000 \\
    VG & 768,536 & 100,406 \\ \midrule
    \textbf{Total} & \textbf{5,247,600} & \textbf{4,126,010} \\ \bottomrule
\end{tabular}\vspace{0.3cm}\caption{Statistics of the 4M pre-training dataset.}\label{tab:dataset_stat}
\end{table}

\end{document}